\documentclass[10pt,twocolumn,letterpaper]{article}

\usepackage{wacv}
\usepackage{times}
\usepackage{epsfig}
\usepackage{makeidx}
\usepackage{graphicx}
\usepackage{amsmath,amssymb} % define this before the line numbering.
\DeclareMathOperator*{\argmax}{\arg\max}   % rbp
\usepackage{hhline}
\usepackage{booktabs}
\usepackage{subcaption}
\usepackage{multirow}
\usepackage{array}
\newcolumntype{L}[1]{>{\raggedright\let\newline\\\arraybackslash\hspace{0pt}}m{#1}}
\newcolumntype{C}[1]{>{\centering\let\newline\\\arraybackslash\hspace{0pt}}m{#1}}
\newcolumntype{R}[1]{>{\raggedleft\let\newline\\\arraybackslash\hspace{0pt}}m{#1}}
\usepackage{paralist}
\wacvfinalcopy % *** Uncomment this line for the final submission

\ifwacvfinal
\def\assignedStartPage{1} % *** Enter the assigned starting page number (instead of 9876)
\fi

\ifwacvfinal
\usepackage[breaklinks=true,bookmarks=false]{hyperref}
\else
\usepackage[pagebackref=true,breaklinks=true,colorlinks,bookmarks=false]{hyperref}
\fi

\ifwacvfinal
\else
\fi

\begin{document}

%%%%%%%%% TITLE
\title{Adaptive-Attentive Geolocalization from few queries: a hybrid approach}

\author{Gabriele Moreno Berton\thanks{The authors equally contributed} , Valerio Paolicelli$^{*}$, Carlo Masone and Barbara Caputo\\
Italian Institute of Technology\\
Turin, Italy\\
{\tt\small [gabriele.berton, valerio.paolicelli, carlo.masone]@iit.it barbara.caputo@polito.it}

}

\maketitle

\begin{abstract}
We address the task of cross-domain visual place recognition, where the goal is to geolocalize a given query image against a labeled gallery, in the case where the query and the gallery belong to different visual domains. To achieve this, we focus on building a domain robust deep network by leveraging over an attention mechanism combined with few-shot unsupervised domain adaptation techniques, where we use a small number of unlabeled target domain images to learn about the target distribution. With our method, we are able to outperform the current state of the art while using two orders of magnitude less target domain images. Finally we propose a new large-scale dataset for cross-domain visual place recognition, called SVOX. The PyTorch code is available at https://github.com/valeriopaolicelli/AdAGeo .
\end{abstract}

\section{INTRODUCTION}
\label{sec:introduction}

In the last decade research on visual place recognition (VPR) has experienced a steady growth, fostered by the availability of large geolocalized image datasets and of smartphones with integrated cameras that make it very easy to capture and share new data.
This growth is confirmed by an increasing number of services that rely on visual place recognition systems, such as 3D reconstruction, consumer photography -"Where did I take these photos?"- and augmented reality.
Moreover, the limitations of localization and orientation systems (e.g. unreliability of GPS signal in urban canyons) make visual place recognition extremely important for the success and scalability of self-driving cars and autonomous robots.

In literature, geolocalization is generally cast as an image retrieval problem. Given a query image as input, the algorithm is tasked to find images that depict the same place from a geotagged dataset, called gallery.
%, where the labels are identifiers of locations (e.g. the name of a landmark \cite{Philbin-2007}, a GPS coordinate \cite{netvlad} or a 6~DoF pose \cite{Sattler-2018}).
%
Most of the recent studies have tried to improve upon this task by using deep convolutional neural networks to extract better representations for the retrieval.  However, they typically consider the case of queries and gallery images belonging to the same domain \cite{kim,multiscale,largescale,apanet}. When the query and gallery images belong to different domains, e.g. due to changes in weather conditions, illumination or season, the performance of such place recognition approaches can be significantly degraded \cite{Sattler-2018,Zaffar-2019}.
Therefore, to make VPR approaches viable in long-term applications we need to explicitly address the cross-domain setting.
%While most of the research in VPR focuses on the case where the queries and the gallery images belong to the same domain \cite{kim,multiscale,largescale,apanet}, or do not explicitly tackle the problem (resulting in poor results when the queries come from other domains), here we focus on the challenging task of cross-domain VPR, in the scenario where  only few unlabeled images from the queries (target) domain are available. \\
%
In this work we tackle this challenge with a novel two-blocks architecture, called AdAGeo, that is aimed at learning robust representations for images for both source (gallery) and target (query) domains. The first block is designed to learn a mapping from the source to the target domain. This mapping is then used to transfer the style of the target domain to the labeled query images of the source domain, as an effective domain-driven data augmentation technique. 
The second block is tasked with producing a representation of the input data that is able to comply with different domains and is suitable for the retrieval task. This is achieved by a combination of an attention module and a domain adaptation module. 
Remarkably, both parts of our architecture only need few unlabeled images from the target domain to be trained. This is of paramount importance to attain a scalable VPR solution, because collecting large amounts of data every time the algorithm needs to be deployed to a different domain is impractical if not infeasible.
To the best of our knowledge, this is the first architecture for few-shot domain adaptation in visual place recognition.
Additionally, for training and validating our method we have built a new large-scale multi-domain dataset, called SVOX (Street View Oxford dataset), that consists of images of Oxford taken from Google Street View (gallery) and queries taken from the Oxford RobotCar dataset \cite{robotcar}. 

\paragraph{Contributions}
\noindent
To summarize, the contributions of our work are:
\begin{itemize}
    \item We present a new dataset, called SVOX, that combines Street View Images (gallery) and queries from the Oxford RobotCar dataset \cite{robotcar}, for the first city-wide multi-domain setting for visual place recognition.
    \item We propose a deep architecture for visual place recognition that combines two orthogonal domain adaptation modules: 
    \begin{inparaenum}[i)]
        \item a generative approach to generate labeled data from the target domain;
        \item a method to produce domain-invariant features. %, thus strengthening cross-domain retrieval.
    \end{inparaenum}
    AdAGeo achieves a significant localization improvement using just few images from the target domain (more than 13\% improvement with 5 images). To the best of our knowledge this is the first hybrid architecture for few-shot domain adaptation in geolocalization.
    \item We propose an attention mechanism through class-specific activation maps, which are used as score maps to weight the features during the image retrieval training and testing processes.
    \item We perform an extensive ablation study as well as comparisons with the current state of the art, demonstrating that our method is able to achieve better performance using just 5 target domain images, while other approaches require hundreds of images.
\end{itemize}
\section{Related works}
\label{sec:relatedworks}
\noindent
In the following we review previous work on VPR and domain adaptation, the two fields closer to our contribution.
%In this paper we tackle the problem of cross-domain VPR, using an attentive deep neural network and domain adaptations techniques to adapt the model to the target domain. Thus, in the next sections we focus on VPR and domain adaptation.\\
\paragraph{Visual Place Recognition.}
Most VPR approaches cast the problem as an image retrieval task \cite{netvlad,spatialPyr,kim,multiscale,deepas,largescale,wang}. This is mostly due to the fact that recent years have seen a huge increase in large scale datasets that cover entire cities or countries, both for research \cite{sped,robotcar,mapillary} and for commercial use (such as Google Street View, Bing Streetside and Apple Maps). 
The increasing availability of datasets has also allowed end-to-end deep learning methods to become dominant in this field, combining deep feature extraction backbones with trainable aggregation modules \cite{netvlad,Gordo-2018} or pooling layers \cite{Radenovic-2019}.
Between the feature extractor and the head, recent architectures for VPR have introduced other modules to improve the retrieval performance.
In particular, following the success of class activation maps (CAM) \cite{cam}, several architectures have implemented some sort of attention module \cite{kim,multiscale,deepas,largescale,apanet} to make the models more robust. 
%Zhu et al \cite{apanet} use a spatial pyramidal pooling block \cite{spatialPyr} to aggregate multi-size regions on the CNN feature maps. 
For what concerns the problem of cross-domain VPR, it has mostly been addressed indirectly and with a limited scope, with approaches that are based on heuristics (e.g. selecting features corresponding to man-made structures \cite{Naseer-2017}), on regions of interest \cite{Chen-2017b}, or tailored for a specific domain shift (e.g. day/night \cite{Torii-2018,Garg-2018}). However, none of these methods allows for generalization.
Only few previous works have explicitly tackled the cross-domain problem. In particular,  \cite{Porav-2018,Anoosheh-2019} both use GANs to replace the query with a synthetic image that depicts the same scene but with the appearance of the source domain.
The authors of \cite{wang} instead use MK-MMD \cite{mkmmd} for domain adaptation and allow the localization of old grayscale photos against a gallery of present-day images. Both source and target datasets are not available at the time of this writing.
While these prior works use either a generative approach or a domain adaptation method, in AdAGeo we combine both solutions and show that there is a benefit to this. Moreover, AdAGeo is truly a few-shot domain adaptation solution that requires as little as 5 unlabeled and not aligned images from the target domain to produce convincing results, whereas \cite{Porav-2018,Anoosheh-2019,wang} need several orders of magnitude more images from the target domain.

\paragraph{Domain adaptation.}
Unsupervised Domain Adaptation attempts to reduce the shift between the source and target distribution of the data by relying only on labelled source data and unlabeled target data.  There are typically two approaches that are used for unsupervised domain adaptation.
The first approach is based on learning a style-transfer transformation to map images from one domain to the other. 
The cross-domain mapping is usually learned through GANs, as in \cite{condGan,augGan}, or autoencoders \cite{vigan}.
The authors of \cite{cyclegan} propose to use a cycle-consistency constraint to learn a meaningful translation, which has since been used in a number of tasks \cite{one_side_domain_mapping,geometry_gan_domain_mapping,cycada,russo17sbadagan}.
The second approach is based on learning domain-invariant features from the data, building on the idea that a good cross-domain representation contains no discriminative information about the origin (i.e. domain) of the input.
This approach was introduced by \cite{grl}, where a domain discriminator network and the gradient reversal layer (GRL) forces the feature extractor to produce domain-invariant representations. 
This method found successful applications in many tasks, such as object detection \cite{grl}, semantic segmentation \cite{Bolte_2019_CVPR_Workshops} and video classification \cite{temporal_attentive_align}.
As an alternative, \cite{SAFN} shows that features with larger norms are more transferable across domains, and proposes to increasingly enlarge the norms of the embedding during training.
In this work we integrate approaches from both kinds in a unique pipeline that only needs few samples from the target domain. We demonstrate via an ablation study that the improvements provided by the two methods are complementary, thus they can be advantageously combined.
\section{Dataset}
\label{sec:3_dataset}
\noindent
In order to address the cross-domain VPR problem we need a dataset that supports different domains between gallery (source) and queries (target). In recent years there have been few VPR datasets that include multiple ambient conditions (weather, seasons, ligthing) \cite{sped,robotcar,mapillary,netvlad,Sunderhauf-2013,Sattler-2018}, however they do not fit our use case due to a limited number of domains \cite{netvlad}, a limited geographical coverage \cite{robotcar,Sattler-2018}, a non dense collection of images \cite{sped} or a non urban setting \cite{Sunderhauf-2013}. 
For this reason we built a new dataset specific for cross-domain VPR in urban setting, called Street View Oxford (SVOX).

%In recent years we have seen a growth in the number of publicly available datasets for urban Visual Place Recognition \cite{sped,robotcar,mapillary,pitts}. Most of them use queries belonging to the same domains as the gallery \cite{pitts}. However, in a real-life application, this might often not be the case. 

% 2) perchè fare cross-domain? esempio
%As an example, consider to build a VPR system in your city, to locate the position of old gray-scale images. You might build a gallery from a widely-available source, such as Google Street View, and use queries coming from a totally different domain. Moreover, it might be cumbersome to acquire a large annotated dataset of target domain images, making source-domain queries an easier choice for training.  

\begin{figure*}
    \centering
    \begin{minipage}{.35\textwidth}
        \begin{subfigure}{\textwidth}
            \includegraphics[width=\textwidth]{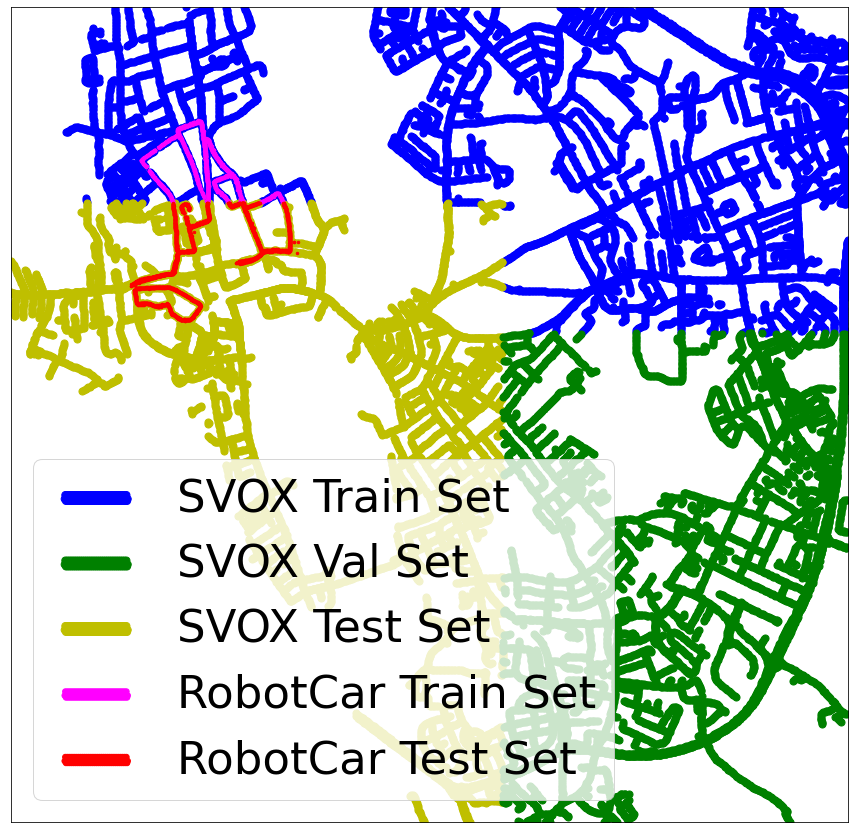}
            \subcaption{}
            \label{fig:map}
        \end{subfigure}
    \end{minipage}
    \begin{minipage}{.2\textwidth}
        \begin{subfigure}{\textwidth}
            \includegraphics[width=\textwidth]{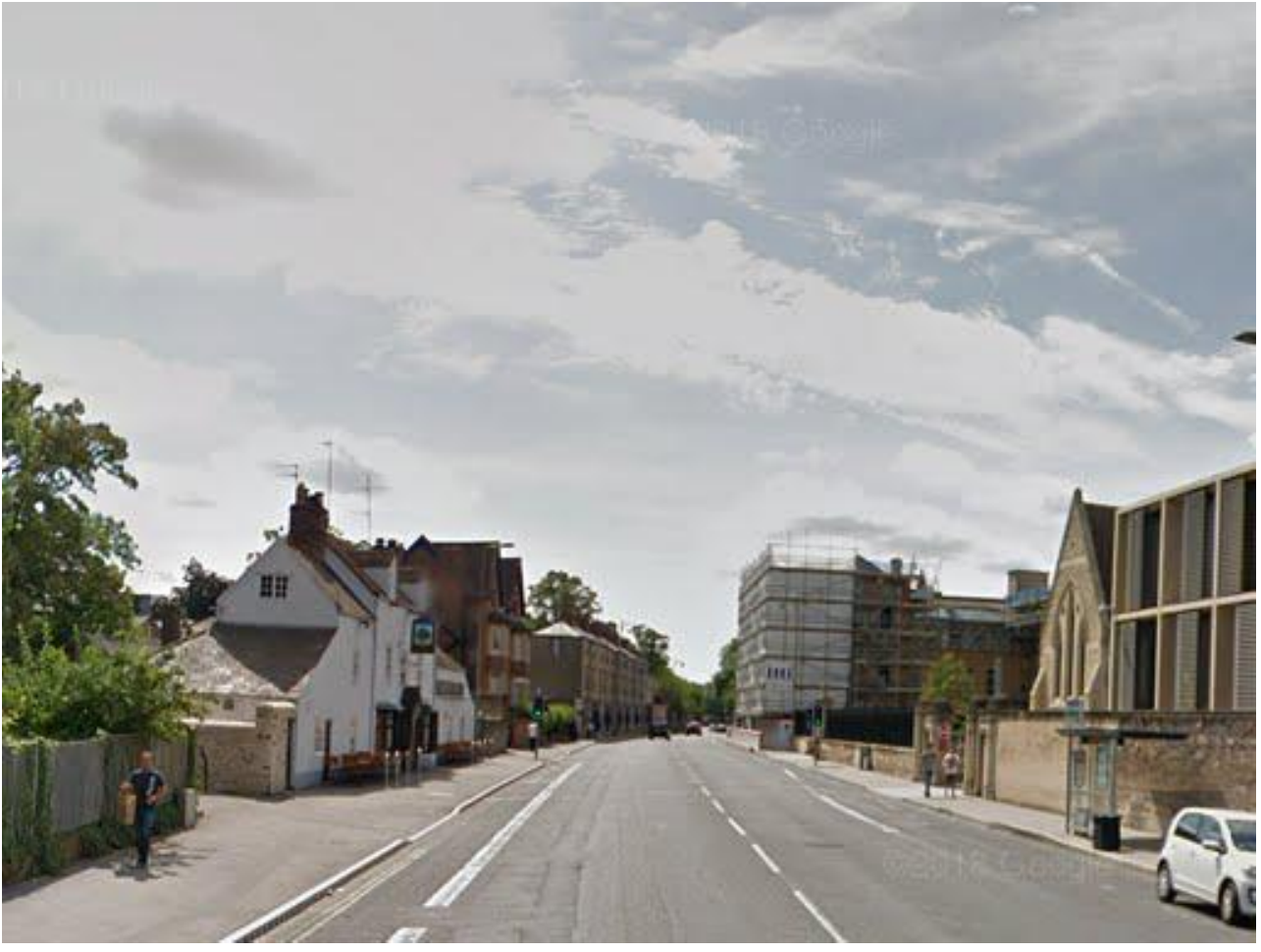} 
            \subcaption{}
            \label{fig:source}
        \end{subfigure}
        \begin{subfigure}{\textwidth}
            \includegraphics[width=\textwidth]{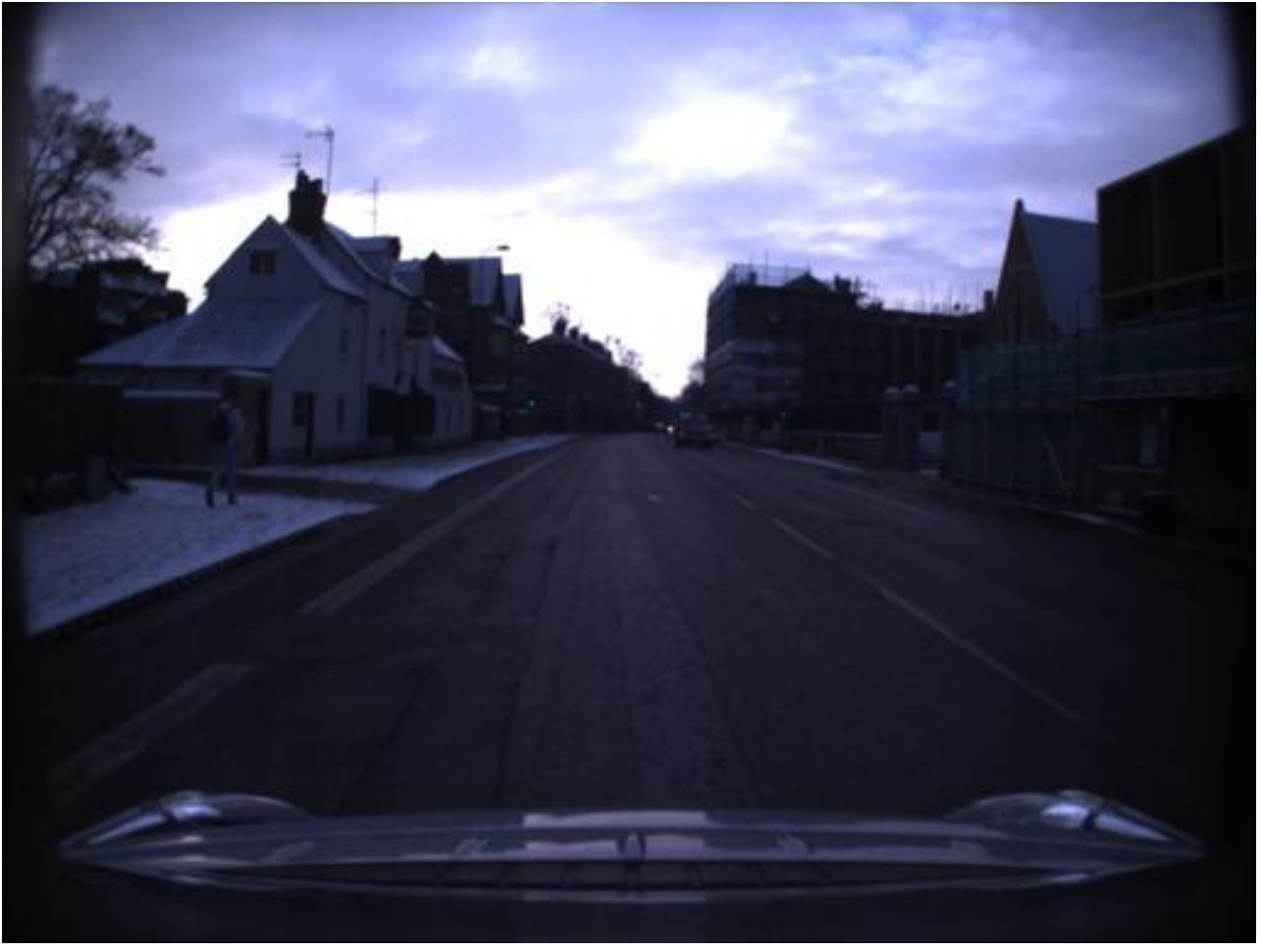}
            \subcaption{}
            \label{fig:snow}
        \end{subfigure}
    \end{minipage}
    \begin{minipage}{.2\textwidth}
        \begin{subfigure}{\textwidth}
            \includegraphics[width=\textwidth]{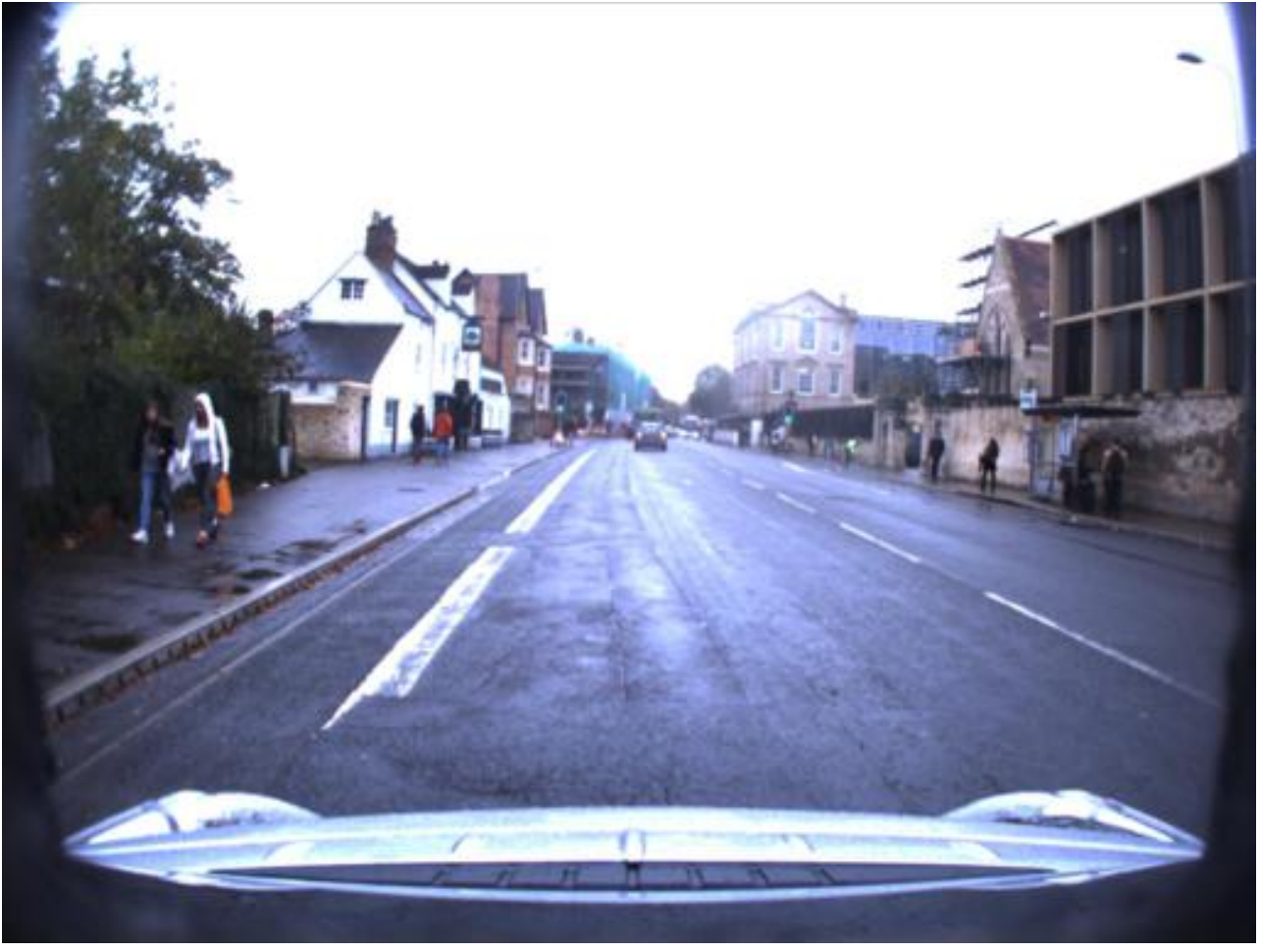} 
            \subcaption{}
            \label{fig:rain}
        \end{subfigure}
        \begin{subfigure}{\textwidth}
            \includegraphics[width=\textwidth]{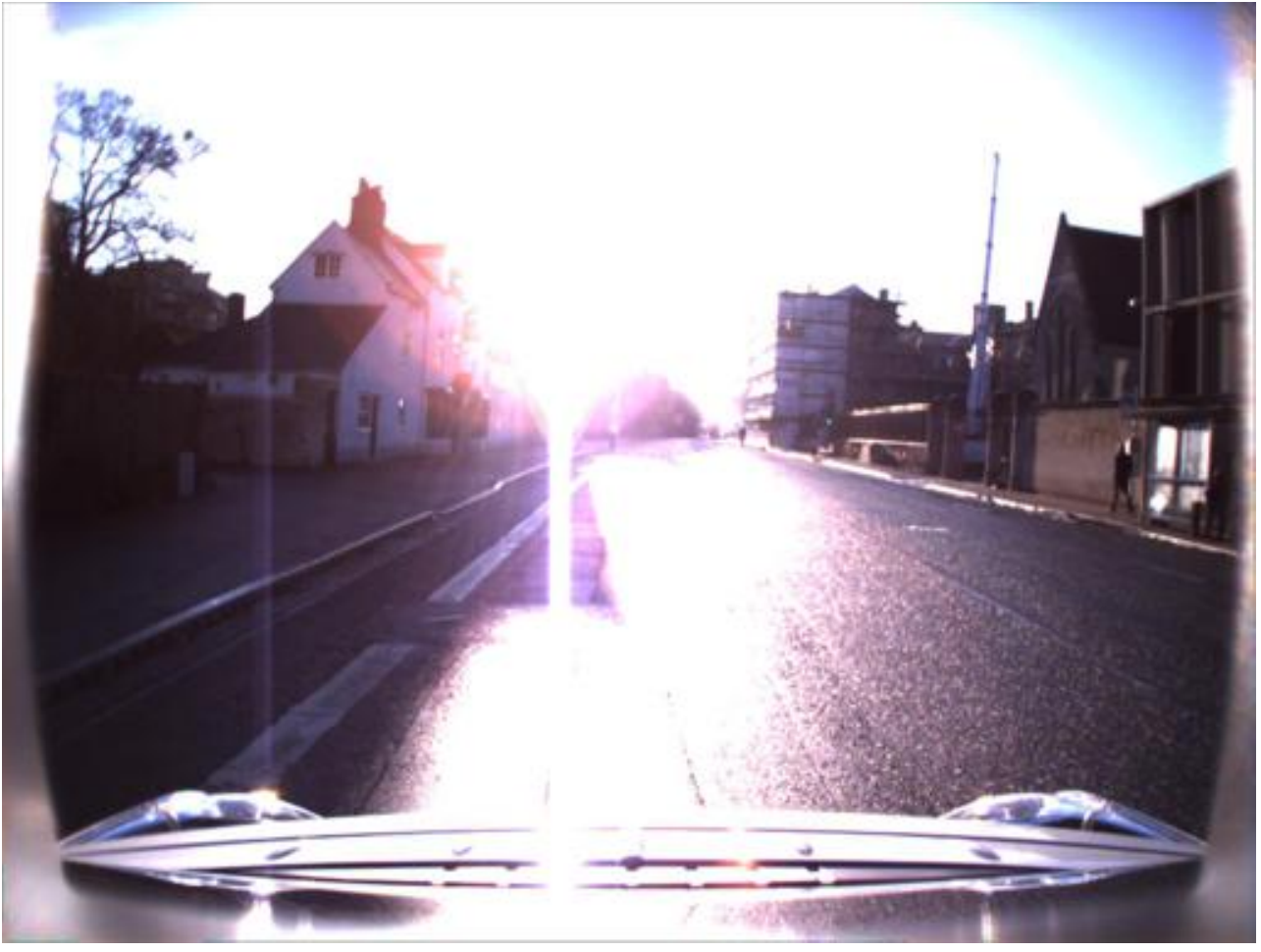} 
            \subcaption{}
            \label{fig:sun}
        \end{subfigure}
    \end{minipage}
    \begin{minipage}{.2\textwidth}
        \begin{subfigure}{\textwidth}
            \includegraphics[width=\textwidth]{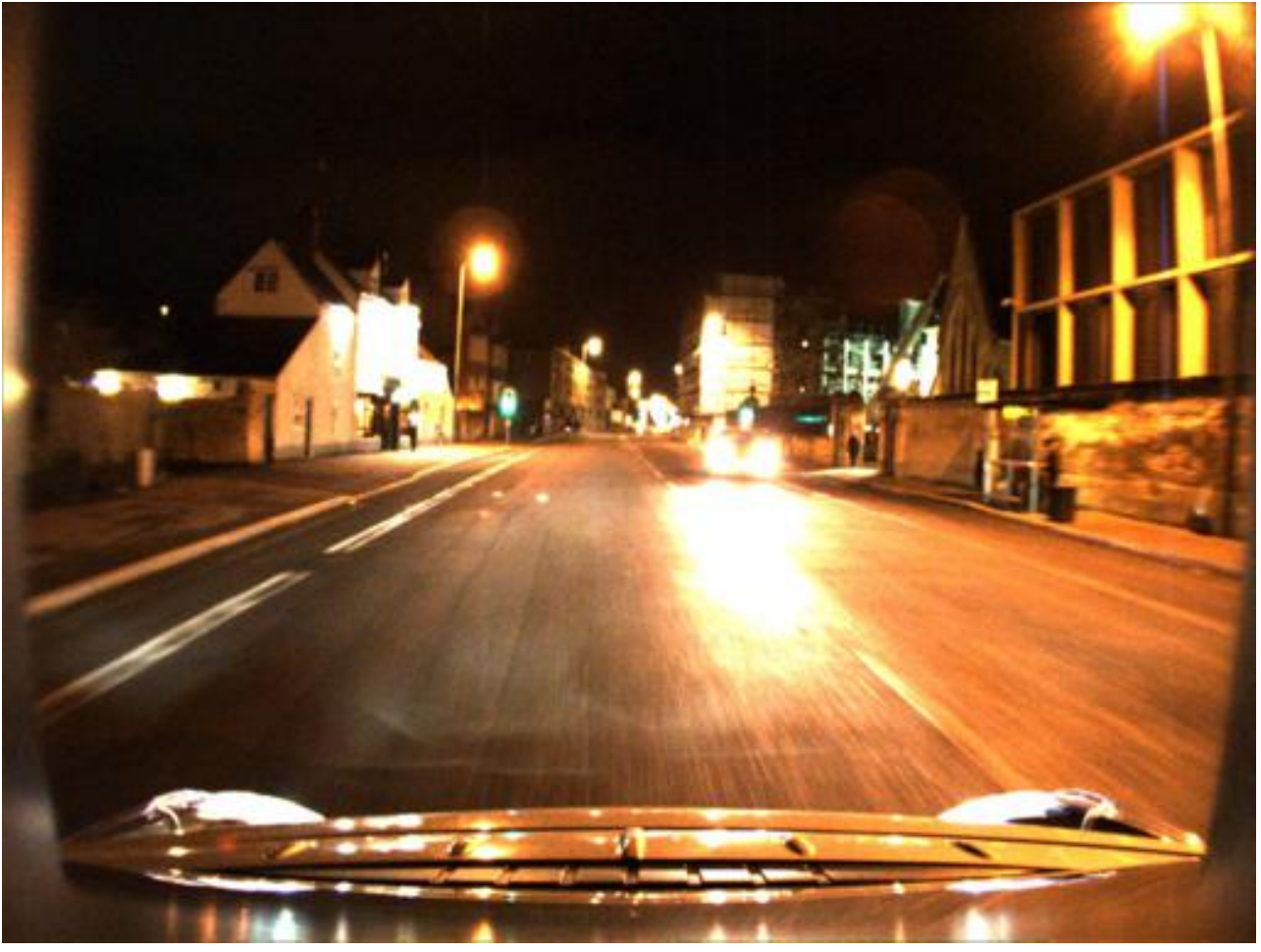}
            \subcaption{}
            \label{fig:night}
        \end{subfigure}
        \begin{subfigure}{\textwidth}
            \includegraphics[width=\textwidth]{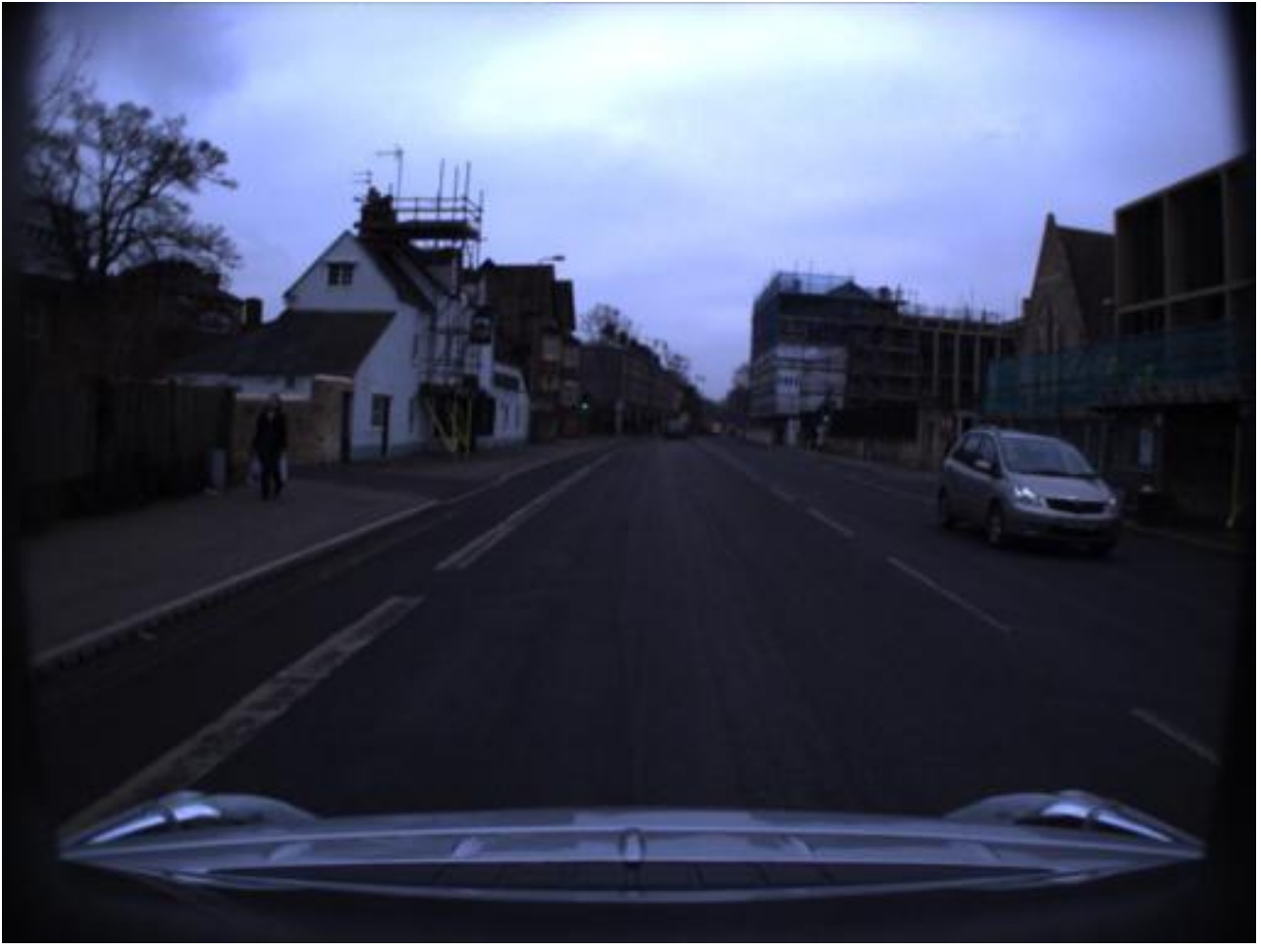}
            \subcaption{}
            \label{fig:overcast}
        \end{subfigure}
    \end{minipage}
    \caption{Combining SVOX and RobotCar datasets: a) shows the areas covered by SVOX and RobotCar \cite{robotcar} on the Oxford city map. b) an example of image from SVOX; c-g) examples from the RobotCar scenarios: respectively Snow, Rain, Sun, Night and Overcast, depicting the same location as image b.}
    \label{fig:dataset_examples}
\end{figure*}

To build SVOX, we used Google Street View to extract images covering a wide area in the city of Oxford. In particular,  we took images from 2012 for the gallery, and images from 2014 as training queries (see Tab. \ref{tab:dataset_counter}), making sure that for each query there is at least one positive sample in the gallery from previous years. We split the dataset in three geographically disjoint subsets, for training, validation and testing. The images collected from Google Street View provide the single source domain. 
Then, to provide different target domains we used samples from the Oxford RobotCar dataset \cite{robotcar} in which images are conveniently tagged according to their weather or lighting conditions. 
For all our experiments we use the 5 domains of Snow, Rain, Sun, Night and Overcast, as defined in the RobotCar dataset. Figures \ref{fig:source}-\ref{fig:overcast} show the differences between the 6 domains. Notice that besides weather, season, and lighting conditions, the RobotCar domains also differ from the source domain for the viewpoint (the hood of the car is visible in the foreground).
Similarly to \cite{Piasco2019LearningSG}, we take one image every 5 meters, in order to avoid using highly redundant data, for example collected when the car was stuck with a red traffic light. This procedure results in roughly 1500 images per domain. %, which are too few for end-to-end training. 
%In this way, we extended RobotCar dataset \cite{robotcar} adding a new and wider scenario, that is far from the ones of \cite{robotcar}.
%
The images collected from the RobotCar dataset are used for domain adaptation and as queries to test the models on different target domains. 
As shown in Fig. \ref{fig:map}, we ensure that for each target query (RobotCar \cite{robotcar}) there is at least one positive sample in the source test gallery (SVOX). Moreover the split is such that the SVOX training data (gallery and query sets) does not overlap RobotCar places (target sets), to avoid possible overfitting.
\begin{table}
    \centering
    \setlength{\tabcolsep}{0.2em}
    \begin{tabular}{r||cc|ccccc} 
        \toprule
        \multicolumn{1}{c||}{\multirow{2}{*}{}} & \multicolumn{2}{c|}{SVOX} & \multicolumn{5}{c}{RobotCar} \\ 
        \multicolumn{1}{c||}{} & Gallery & Queries & Snow & Rain & Sun & Night & Overcast \\ 
        \hline
        Train & 22232 & 11294 & 750 & 714 & 712 & 702 & 705 \\
        Val   & 17226 & 14698 &  -  &  -  &  -  &  -  &  -  \\
        Test  & 17166 & 14278 & 937 & 870 & 854 & 823 & 872 \\
        \bottomrule
    \end{tabular}
    \caption{Sizes of SVOX dataset and Oxford RobotCar \cite{robotcar} from 5 different scenarios}
    \label{tab:dataset_counter}
\end{table}

Further details about the procedure implemented to collect  SVOX are provided in the supplementary material.

\section{Method}
\label{sec:4_method}
\begin{figure*}
    \centering
    \includegraphics[width=\linewidth]{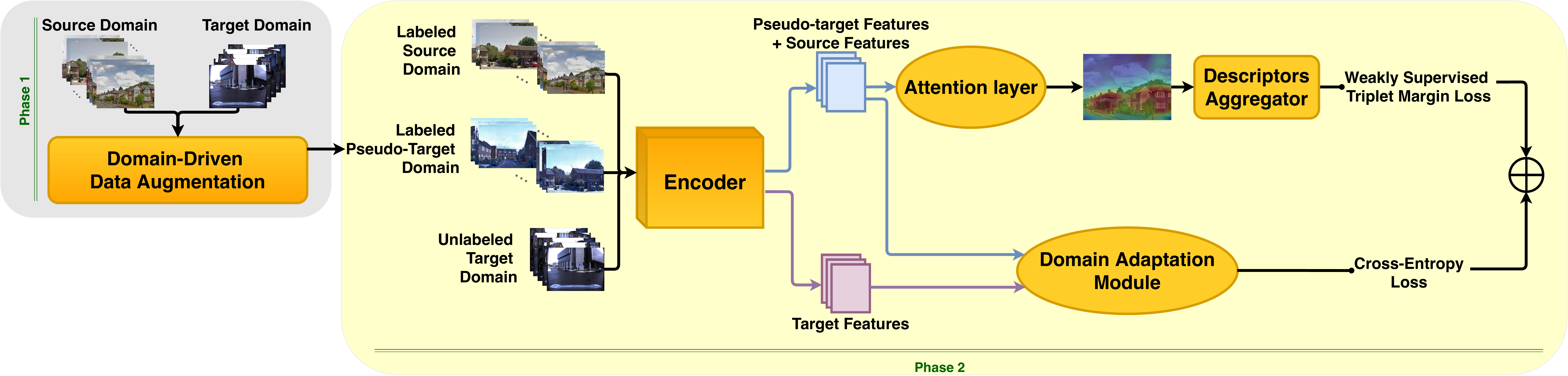}
    \caption{The proposed AdAGeo architecture: training is performed in two phases: during phase 1 the domain-driven data augmentation learns a transformation from source to target domain from just 5 target images. The transformation is then used to generate labeled pseudo-target images. Phase 2 is tasked with the actual geolocalization task, leveraging the generated pseudo-target images, an attention layer and a domain adaptation module.}
    \label{fig:ch3_arch}
\end{figure*}

\noindent
In this section we present AdAGeo, our method for domain adaptive and attentive visual place recognition. The architecture is composed of two parts (Fig. \ref{fig:ch3_arch}), which are trained separately. 
The first one is a few-shot domain-driven data augmentation (DDDA) module (Sec. \ref{subsec:4.1_few-shotDDDA}). By using just few images from the target domain, this module is able to effectively transfer their style to the source domain images. In this way we can use these labeled augmented images to make the VPR model robust to the target domain.
The second block is made of a CNN encoder, which extracts features for the domain adaptation (DA) module (Sec. \ref{subsec:4.3_grl}), and for the attention (Att) module (Sec. \ref{subsec:4.2_attention}) followed by a descriptors aggregator (Sec. \ref{subsec:4.3_netvlad}), which builds robust attentive embeddings for each image.\\
As shown in Fig. \ref{fig:ch3_arch}, during the phase 2, the network receives the SVOX gallery set as retrieval gallery, the SVOX query set and the related pseudo-target images as queries to perform the main task, while the unsupervised domain adaptation (DA) task is computed over SVOX, pseudo-target and just a few target images.

\subsection{Few-shot domain-driven data augmentation}
\label{subsec:4.1_few-shotDDDA}
\noindent
%questa prima parte di presentazione di DA è molto simile a https://arxiv.org/pdf/1707.01217.pdf
In unsupervised domain adaptation we have a labeled source dataset $X^{s} = \{(x_{i}^{s}, y_{i}^{s})\} _{i=1}^{n^{s}}$ made of $n^{s}$ samples (comprising gallery and queries) from source domain $D_{s}$, and an unlabeled target dataset $X^{t} = \{(x_{j}^{t})\} _{j=1}^{n^{t}}$ made of $n^{t}$ samples from target domain $D_{t}$.
The goal of our few-shot domain-driven data augmentation is to learn a mapping from $D_{s}$ to $D_{t}$, in the case where $n^{t}$ is small (results of of experiments with different values of $n^{t}$ are later shown in Fig. \ref{fig:exp_ntarget}). This mapping is used as data augmentation for the training queries, to generate labeled target domain queries, and to ultimately make the image retrieval model more robust to the domain shift. 
We take inspiration from \cite{biost}, which proposes an approach for the related problem of learning a bi-directional mapping between two domains, for which they only have one sample belonging to $D_t$. The idea is to use an architecture made of two parallel autoencoders, one for each domain. 
Let us call $Ae_{S}$ and $Ae_{T}$ the two autoencoders, where $Ae_{S}(x) = Dec_{S} (Enc_{S} (x))$ and $Ae_{T}(x) = Dec_{T} (Enc_{T} (x))$, with $Enc_{S}$ and $Enc_{T}$ denoting encoders and $Dec_{S}$ and $Dec_{T}$ decoders.
The goal is to minimize the distance between the distributions of the latent spaces of the two autoencoders, forcing the encoders to produce domain-invariant embeddings, while at the same time each decoder should be able to translate the embeddings to an image in its own domain.
This is achieved by minimizing a reconstruction loss on both autoencoders: 
\begin{equation}
    L_{REC} = \sum_{s \in S} \Vert Ae_{S}(s) ) - s \Vert _{1} + \sum_{t \in T} \Vert Ae_{T}(t) ) - t \Vert _{1}
\end{equation}
as well as cycle-consistency losses:
\begin{equation}
\begin{split}
    L_{sts-cycle} = \sum_{s \in S} \Vert Dec_{S} (\overline{Enc_{T}} (\overline{Dec_{T}} (Enc_{S} (s)))) - s \Vert _{1}\\
    L_{tst-cycle} = \sum_{t \in T} \Vert Dec_{T} (\overline{Enc_{S}} (\overline{Dec_{S}} (Enc_{T} (t)))) - t \Vert _{1}
\end{split}
\end{equation}
where the bar above a module means that its weights are frozen during backpropagation of this loss. Moreover, it is important that the embeddings approximate a Gaussian distribution, which helps the two domains to better align, and can be achieved through a variational loss on both encoders:
\begin{equation}
\begin{split}
    L_{V Enc_{S}} = \sum_{s \in S} KL(\{ Enc_{S}(s) | s \in S \} \Vert \hspace{1pt} \mathcal{N}(0,I))\\
    L_{V Enc_{T}} = \sum_{t \in T} KL(\{ Enc_{T}(t) | t \in T \} \Vert \hspace{1pt} \mathcal{N}(0,I))
\end{split}
\end{equation}
We can then compute the final loss as:
\begin{equation}
\begin{split}
    L_{final} = L_{REC} + L_{sts-cycle} + L_{tst-cycle} + \\ 0.001 L_{V Enc_{S}} + 0.001 L_{V Enc_{T}}
\end{split}
\end{equation}
Once the training process is finished, it is possible to generate new images from the source domain dataset $X^{s}$, by translating them into the target domain $D_{t}$.
We therefore generate a new pseudo-target dataset $X^{pt} = \{(x_{i}^{pt}, y_{i}^{pt})\} _{i=1}^{n^{pt}}$ where $n^{pt} = n^{s}$, $x_{i}^{pt} = Dec_{T}(Enc_{S}(x_{i}^{s}))$ and $y_{i}^{pt} = y_{i}^{s}$ for all $i \in \{1, 2... n^{pt}\}$. We call this pseudo-target because its domain $D_{pt} \approx D_{t}$. The creation of the pseudo-target dataset is a data augmentation technique performed only once, offline, in order to speed up the training of the second part of the architecture.
\label{sec:3_style_transf}
\begin{figure}
    \centering
    \begin{minipage}{.14\textwidth}
        \begin{subfigure}{\textwidth}
            \includegraphics[width=\textwidth]{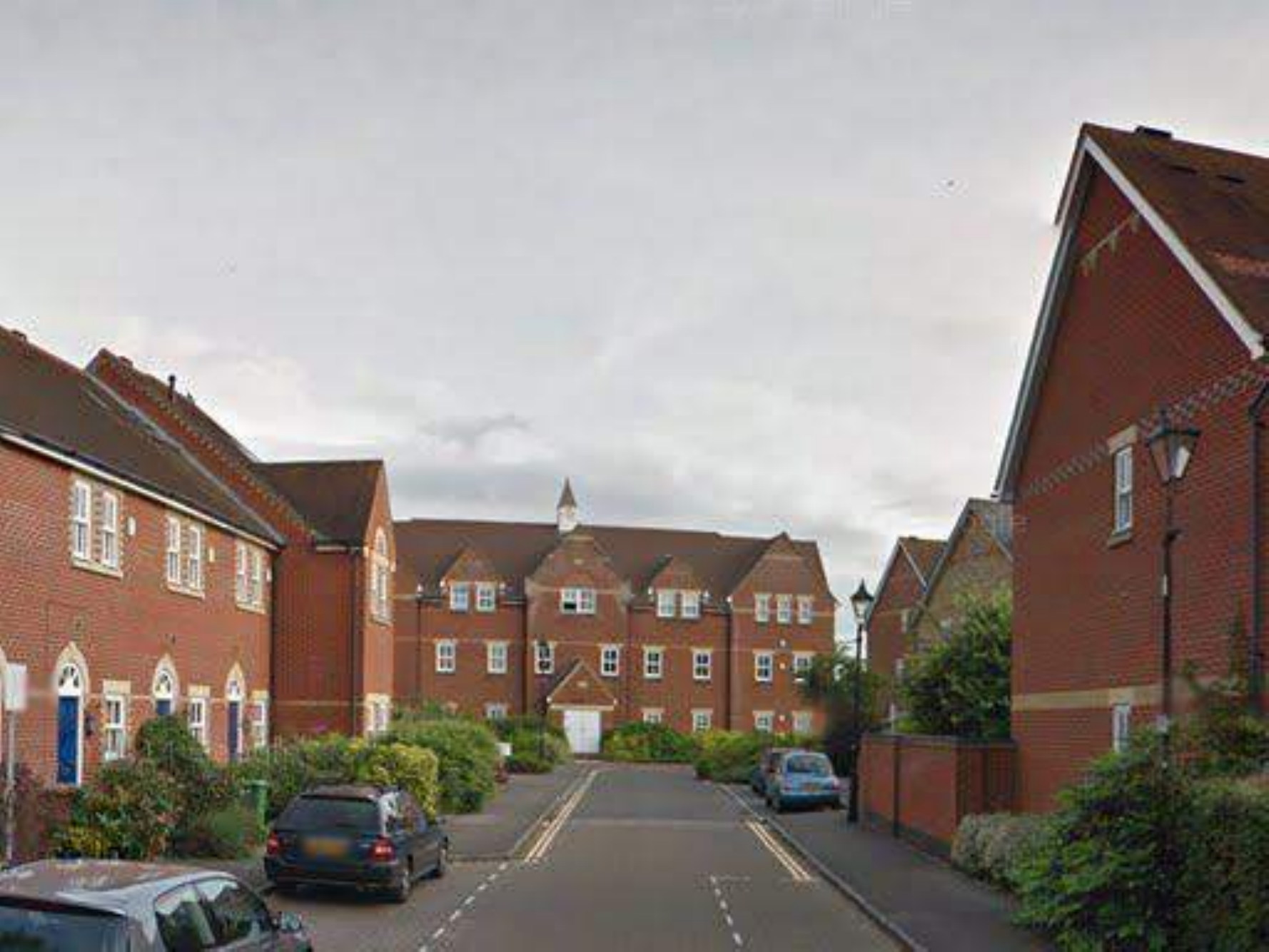} \subcaption{Original}
        \end{subfigure}
    \end{minipage}
    \begin{minipage}{.14\textwidth}
        \begin{subfigure}{\textwidth}
            \includegraphics[width=\textwidth]{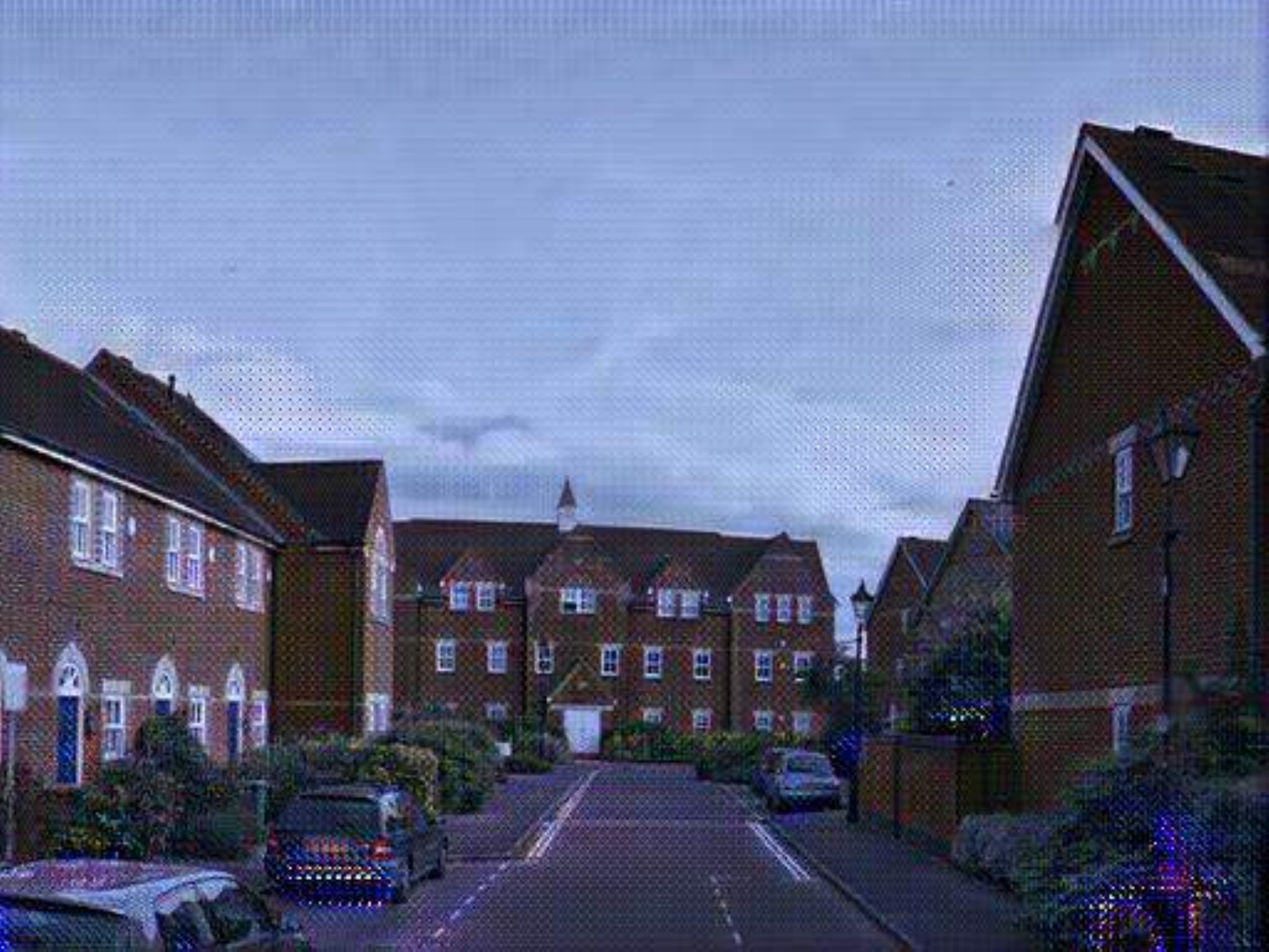} \subcaption{Snow}
        \end{subfigure}
    \end{minipage}
    \begin{minipage}{.14\textwidth}
        \begin{subfigure}{\textwidth}
            \includegraphics[width=\textwidth]{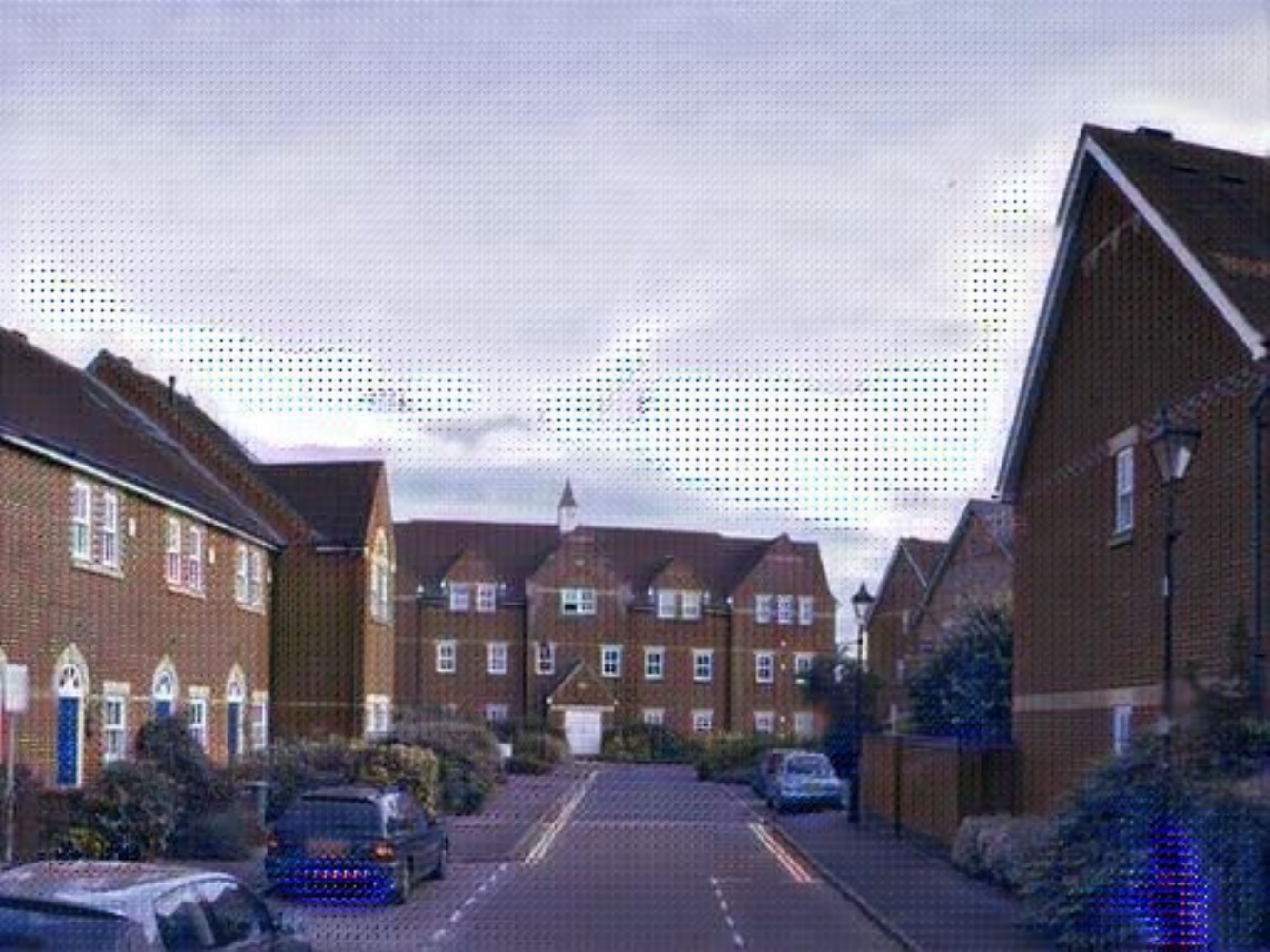} \subcaption{Rain}
        \end{subfigure}
    \end{minipage}
    \begin{minipage}{.14\textwidth}
        \begin{subfigure}{\textwidth}
            \includegraphics[width=\textwidth]{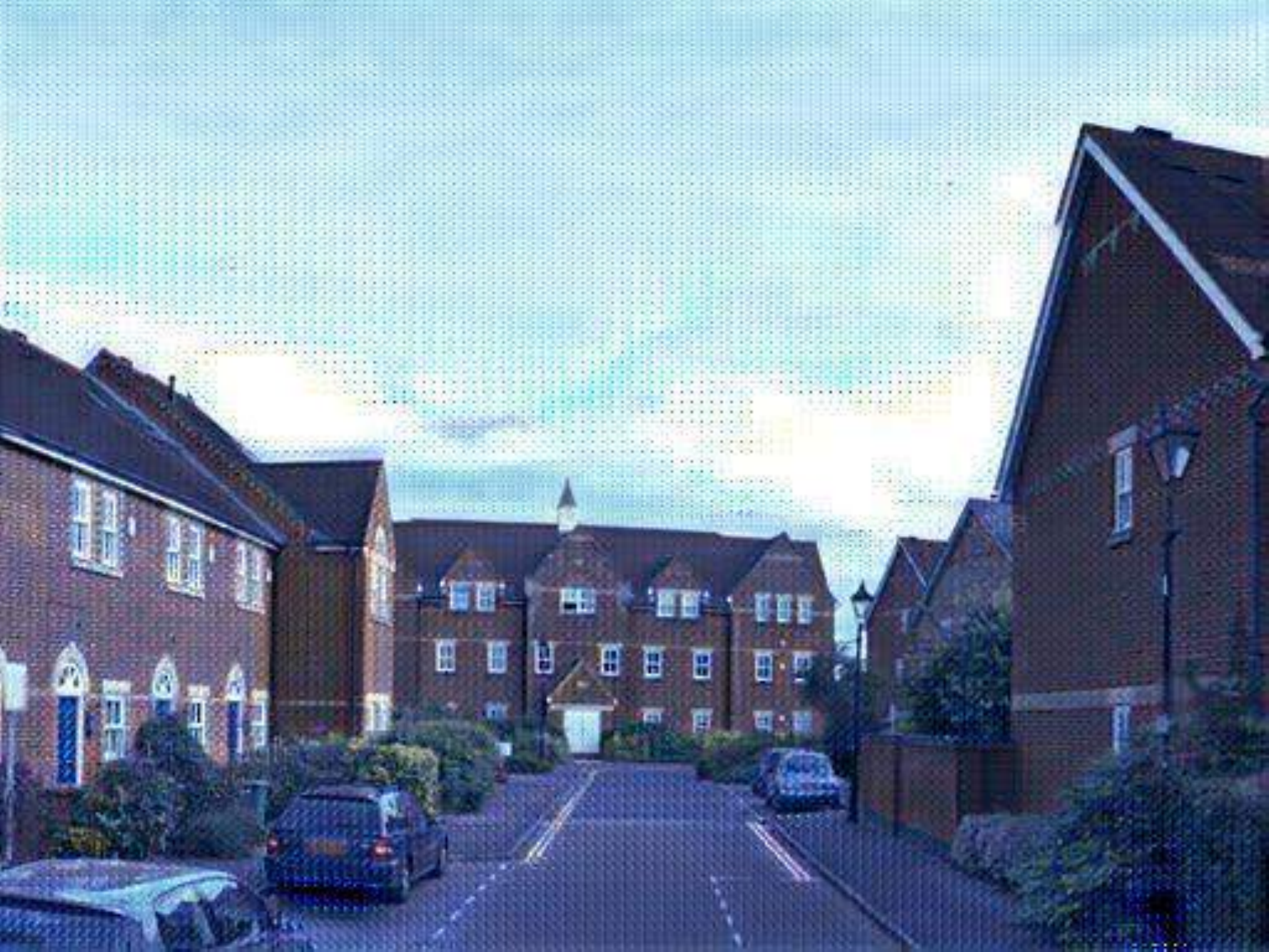} \subcaption{Sun}
        \end{subfigure}
    \end{minipage}
    \begin{minipage}{.14\textwidth}
        \begin{subfigure}{\textwidth}
            \includegraphics[width=\textwidth]{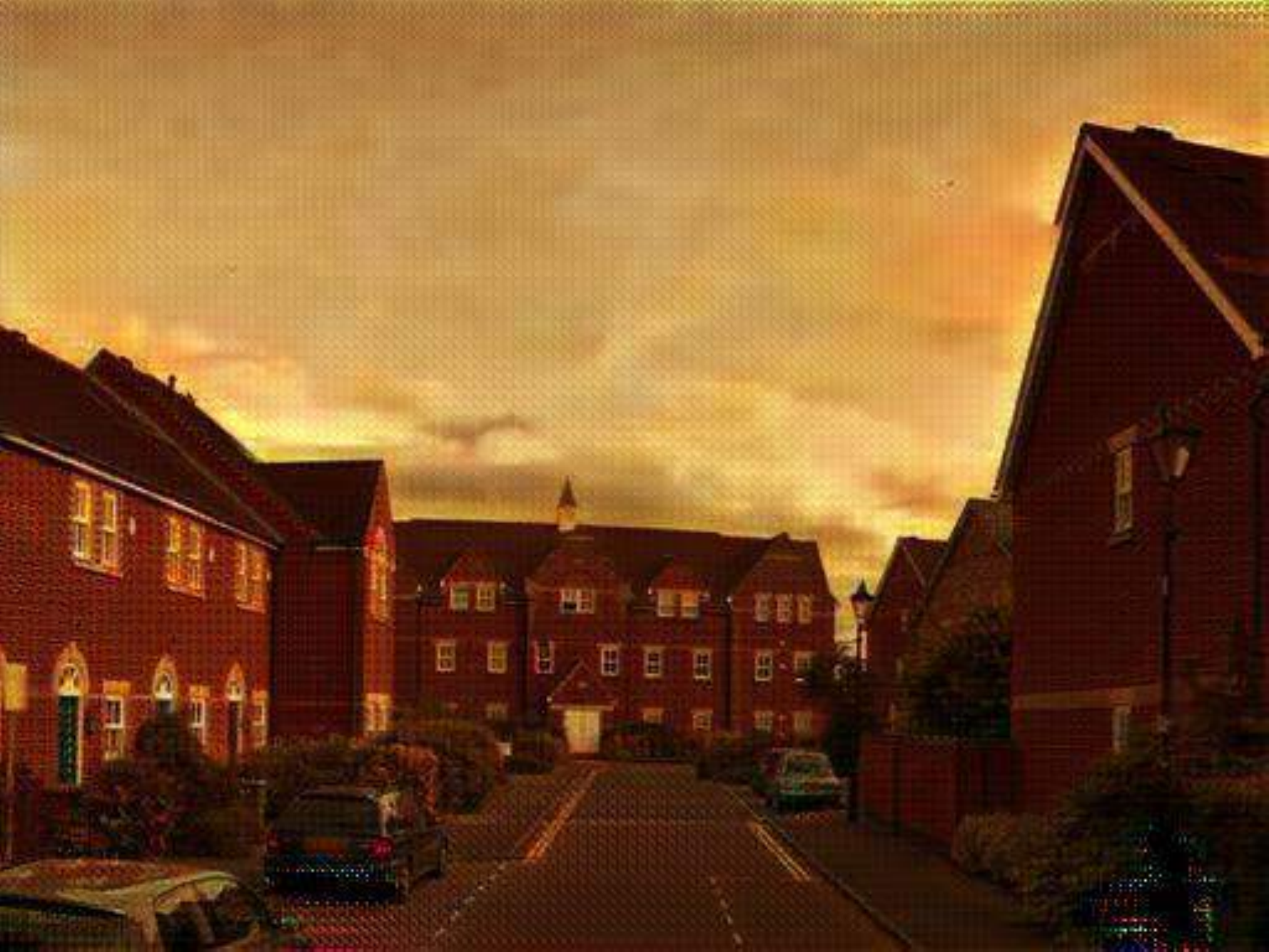} \subcaption{Night}
        \end{subfigure}
    \end{minipage}
    \begin{minipage}{.14\textwidth}
        \begin{subfigure}{\textwidth}
            \includegraphics[width=\textwidth]{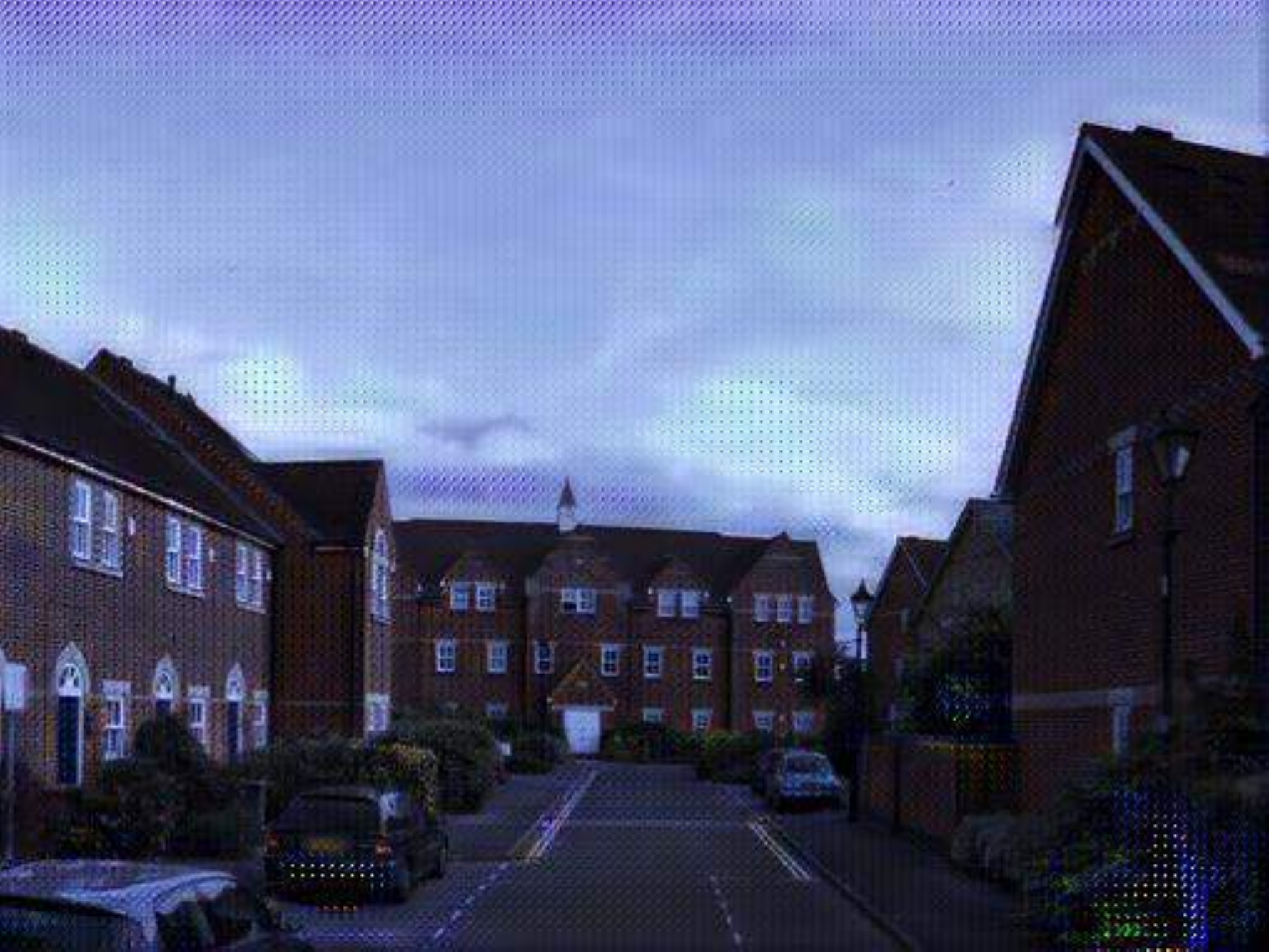} \subcaption{Overcast}
        \end{subfigure}
    \end{minipage}
    \caption{An image from the SVOX dataset (a) and the 5 generated pseudo-target images over the 5 domains of RobotCar. Although the generated images present visible artifacts, this step is essential for cross-domain robust geolocalization (see Tab. \ref{tab_ablation}).}
    \label{fig:biost_style_transfer}
\end{figure}

\subsection{Attention mechanism}
\label{subsec:4.2_attention}
\noindent
In order to highlight the most important features' areas for the retrieval task, we introduced an attention layer after the encoder.
To this purpose, we took inspiration from the class activation map (CAM) paper \cite{cam} which tries to focus on discriminative image areas that are the most useful to produce the class output in the image classification task, exploiting the final average pooling layer present in recent networks such as the ResNet \cite{resnet}.
Let us consider for a given image of dimension $3 \times H \times W$ the extracted feature representation $f$ of shape $D \times H_{1} \times W_{1}$ where $D$ is the number of kernels from the last convolutional layer in the encoder. Furthermore, consider also the backbone classifier block, which contains a fully connected layer with $D \times C$ weights $w_{cd}$ with $d$ values respectively for each class $c$. The attention map $AM_c$ for a given class $c$ is obtained by the following linear combination:
\begin{equation}
    AM_{c} = \sigma (\sum_{d} \hspace{1pt} f_d \cdot w_{cd})
\end{equation}
where $\sigma$ is the softmax function and whose result has dimension $H_{1} \times W_{1}$.\\
Finally, $AM_{c}$ is upsampled to $H \times W$ and is applied over the input image,  to visualize the most relevant regions for that class c.

\begin{figure}
    \centering
    \begin{minipage}{.12\textwidth}
        \begin{subfigure}{\textwidth}
            \includegraphics[width=\textwidth]{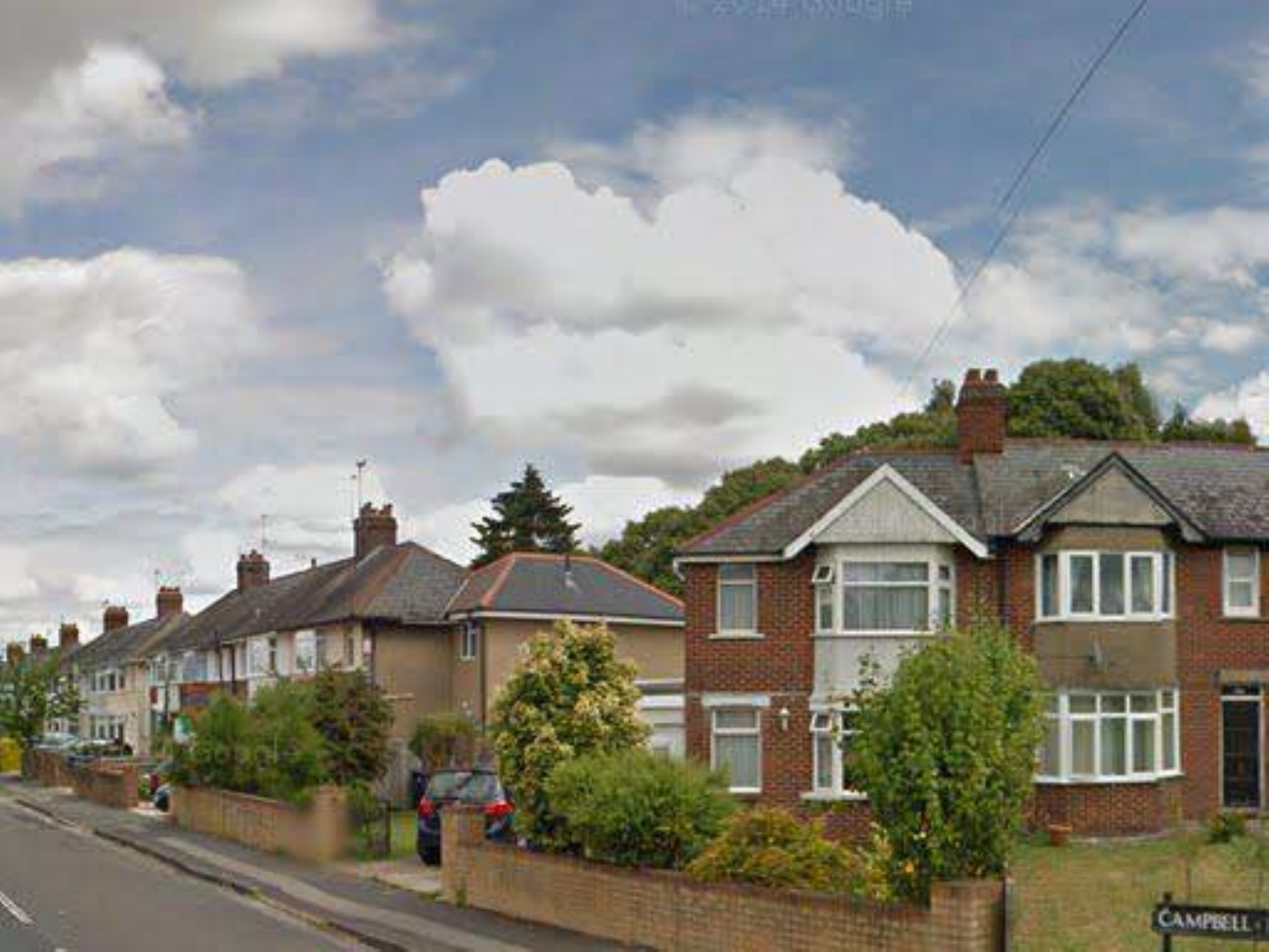} 
        \end{subfigure}
        \begin{subfigure}{\textwidth}
            \includegraphics[width=\textwidth]{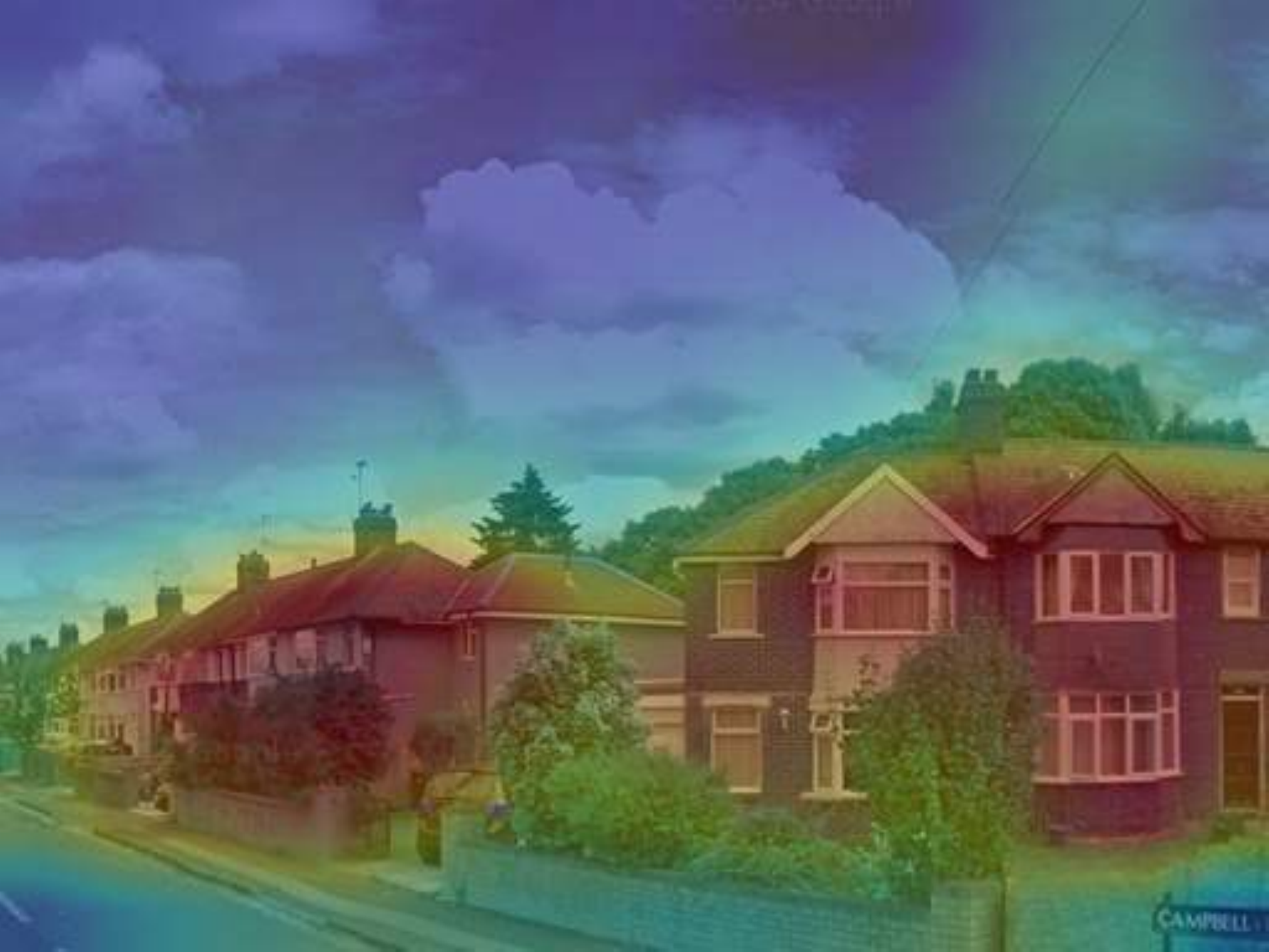} \subcaption{}
        \end{subfigure}
    \end{minipage}
    \begin{minipage}{.12\textwidth}
        \begin{subfigure}{\textwidth}
            \includegraphics[width=\textwidth]{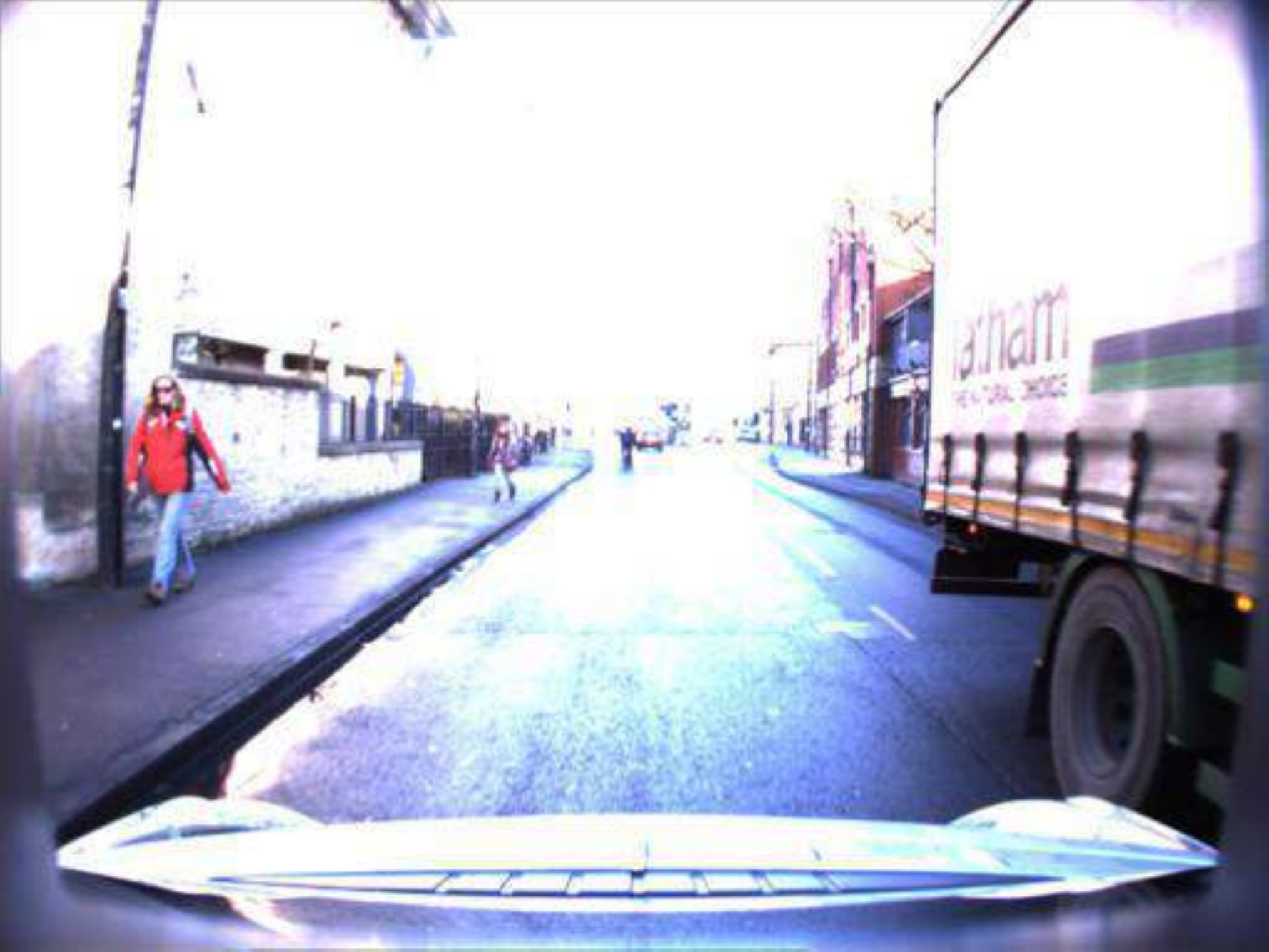} 
        \end{subfigure}
        \begin{subfigure}{\textwidth}
            \includegraphics[width=\textwidth]{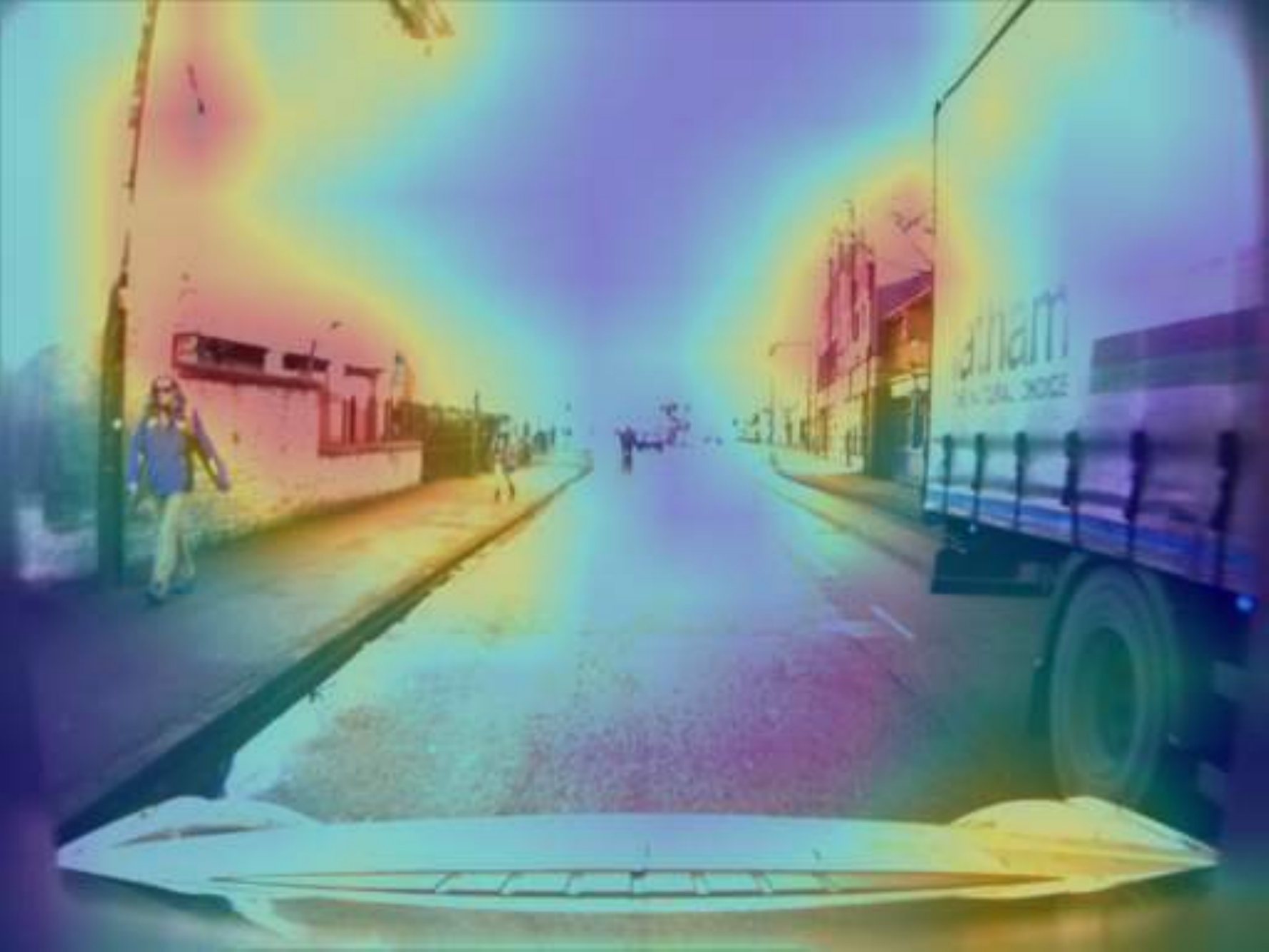} \subcaption{}
        \end{subfigure}
    \end{minipage}
    \begin{minipage}{.12\textwidth}
        \begin{subfigure}{\textwidth}
            \includegraphics[width=\textwidth]{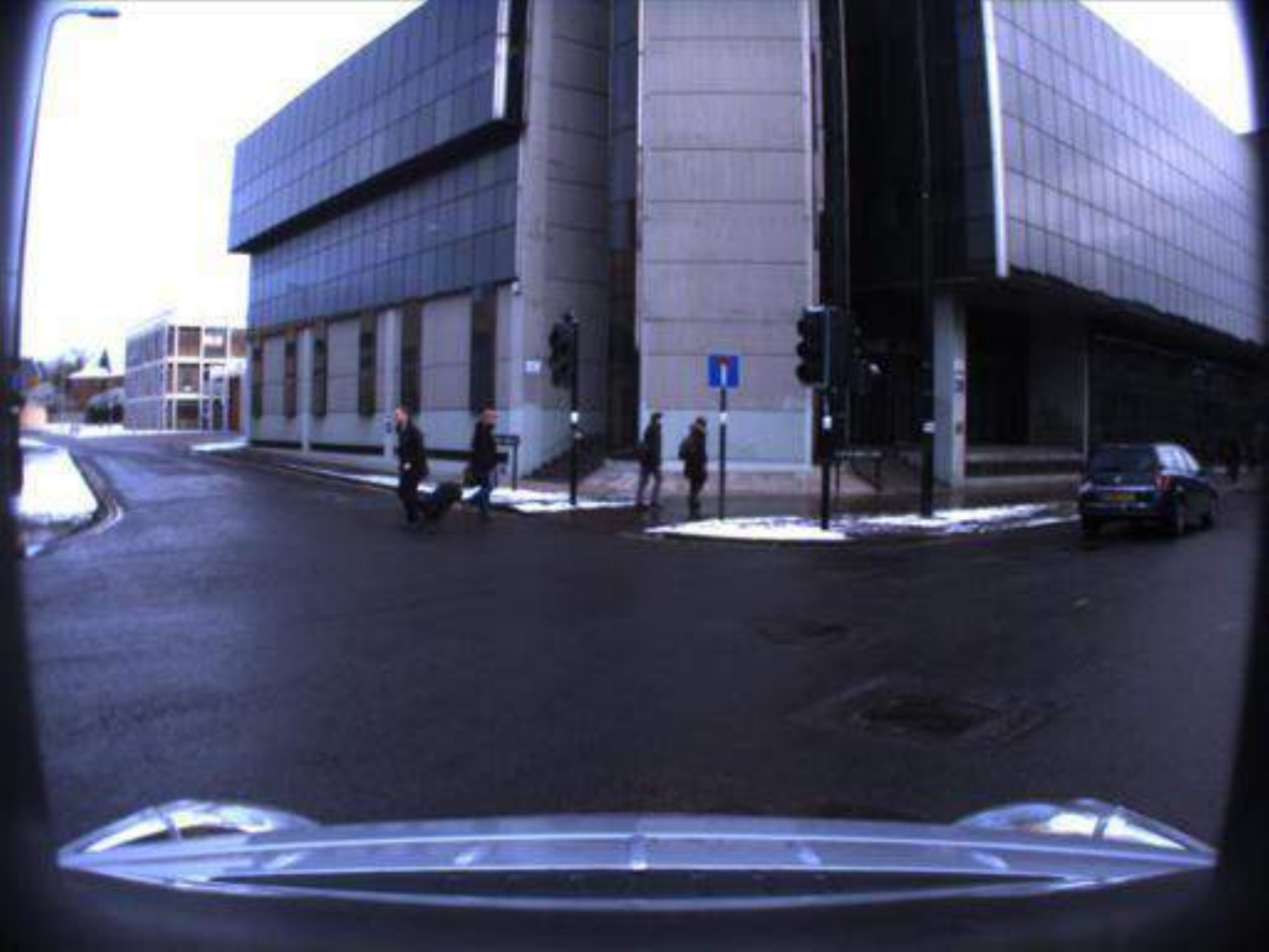} 
        \end{subfigure}
        \begin{subfigure}{\textwidth}
            \includegraphics[width=\textwidth]{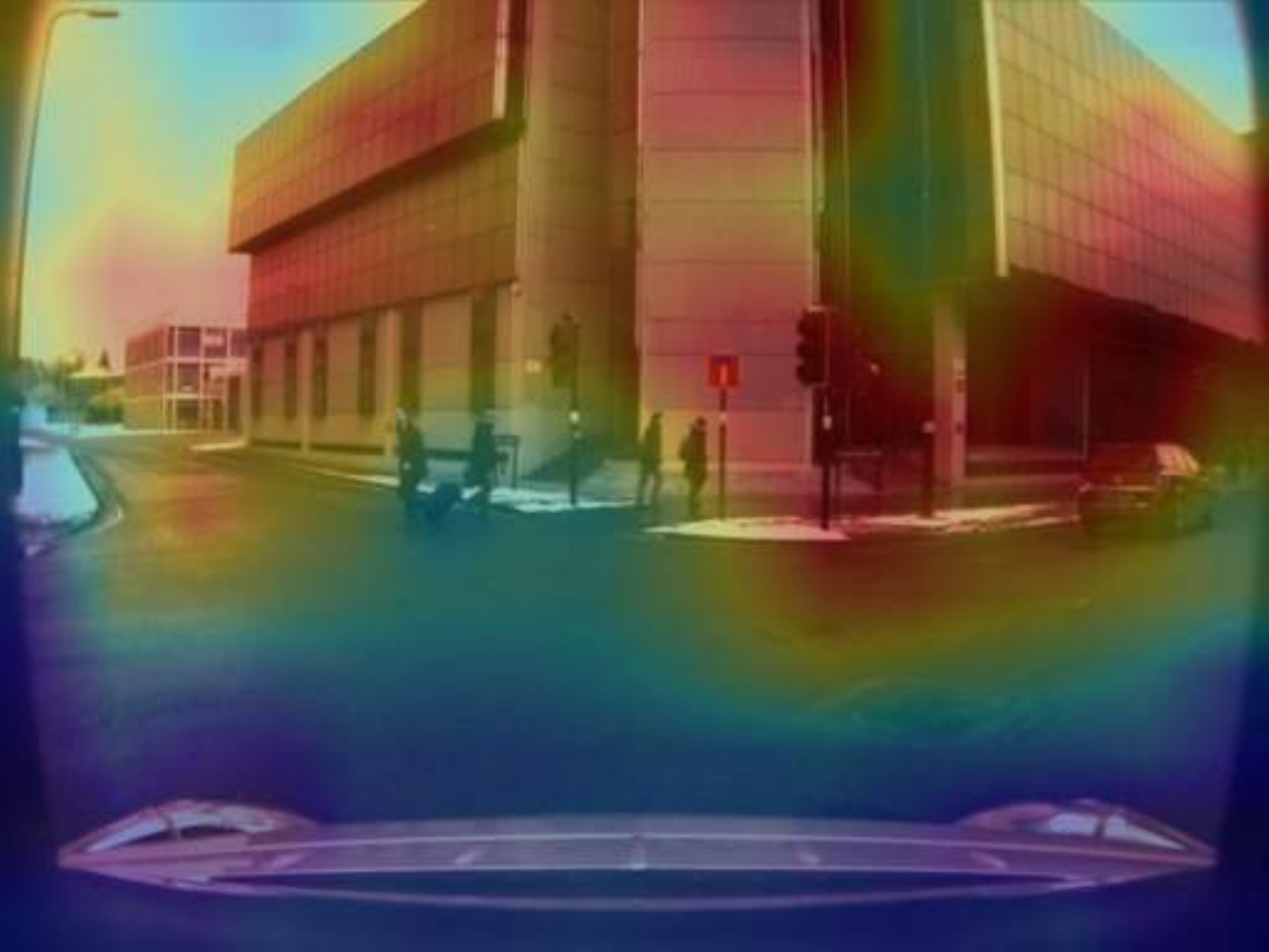} \subcaption{}
        \end{subfigure}
    \end{minipage}\\
    \begin{minipage}{.12\textwidth}
        \begin{subfigure}{\textwidth}
            \includegraphics[width=\textwidth]{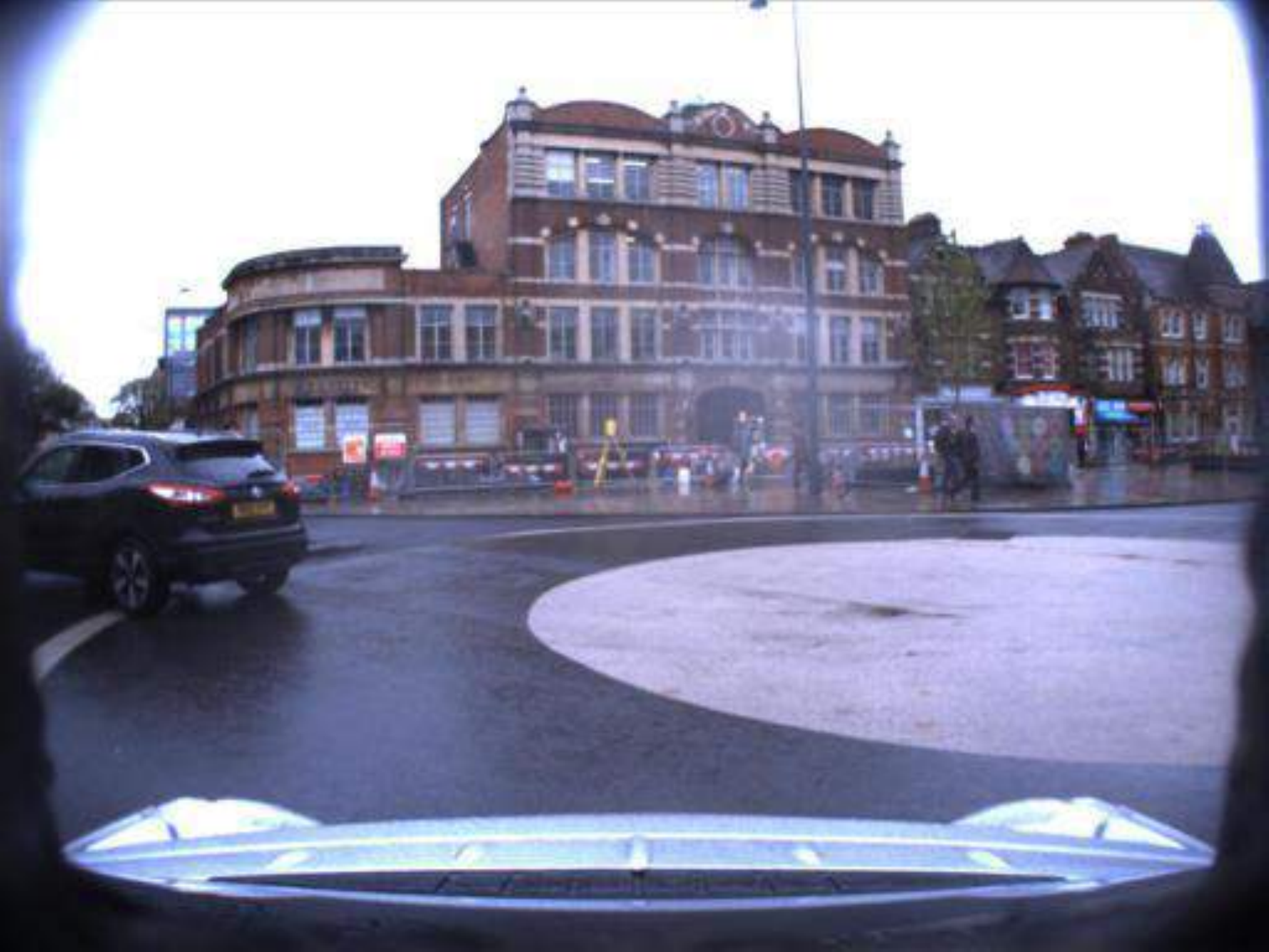} 
        \end{subfigure}
        \begin{subfigure}{\textwidth}
            \includegraphics[width=\textwidth]{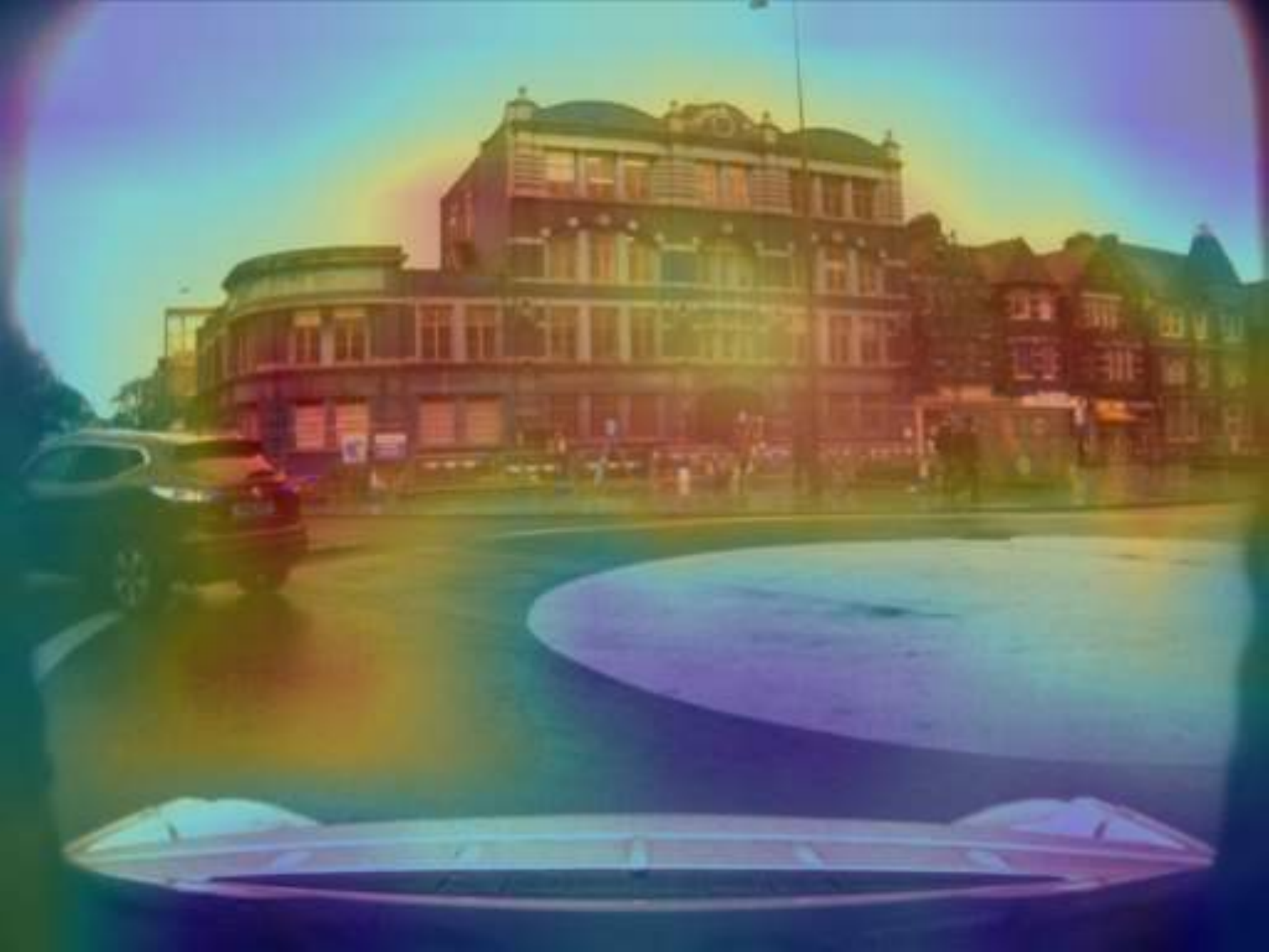} \subcaption{}
        \end{subfigure}
    \end{minipage}
    \begin{minipage}{.12\textwidth}
        \begin{subfigure}{\textwidth}
            \includegraphics[width=\textwidth]{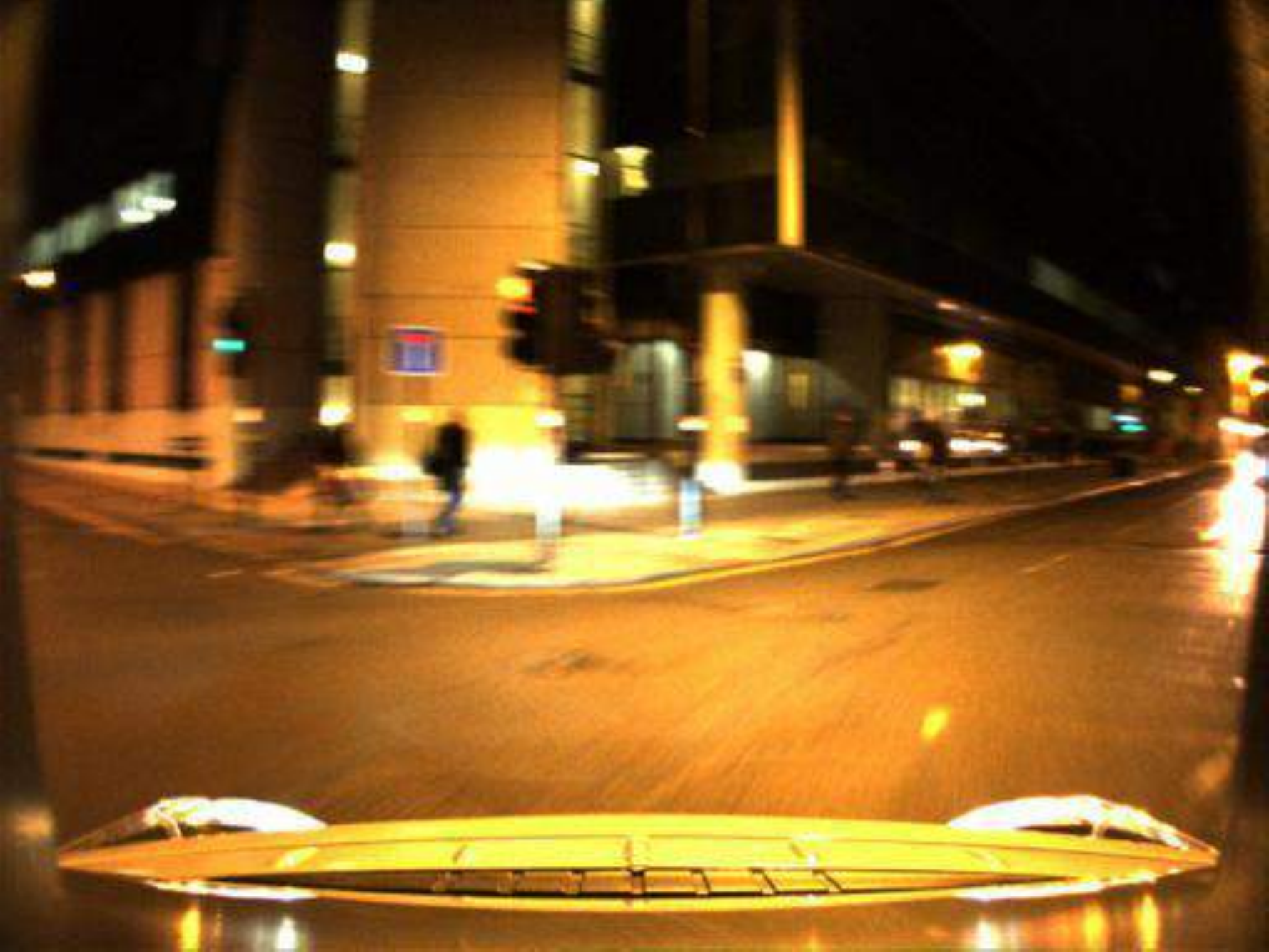} 
        \end{subfigure}
        \begin{subfigure}{\textwidth}
            \includegraphics[width=\textwidth]{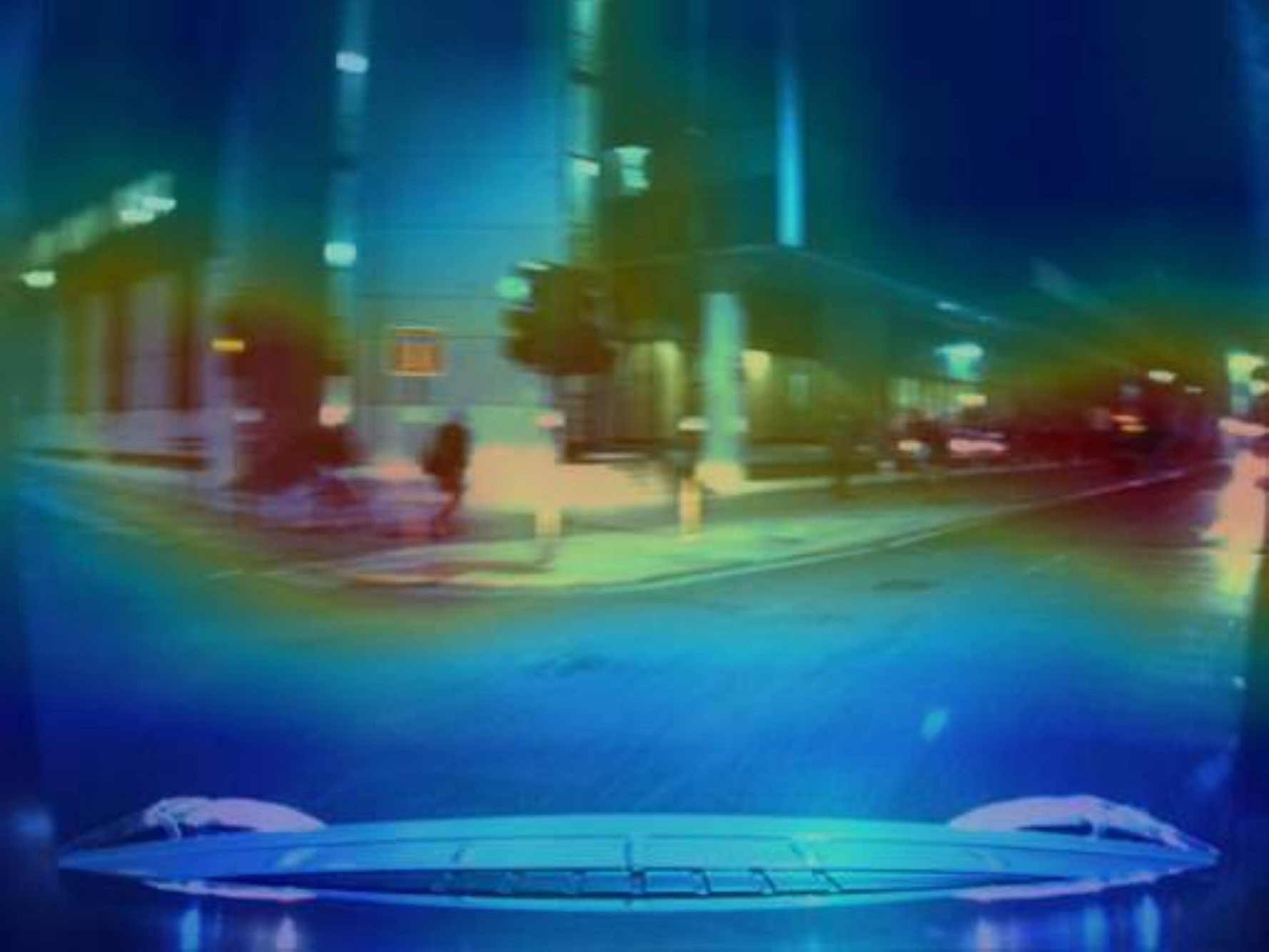} \subcaption{}
        \end{subfigure}
    \end{minipage}
    \begin{minipage}{.12\textwidth}
        \begin{subfigure}{\textwidth}
            \includegraphics[width=\textwidth]{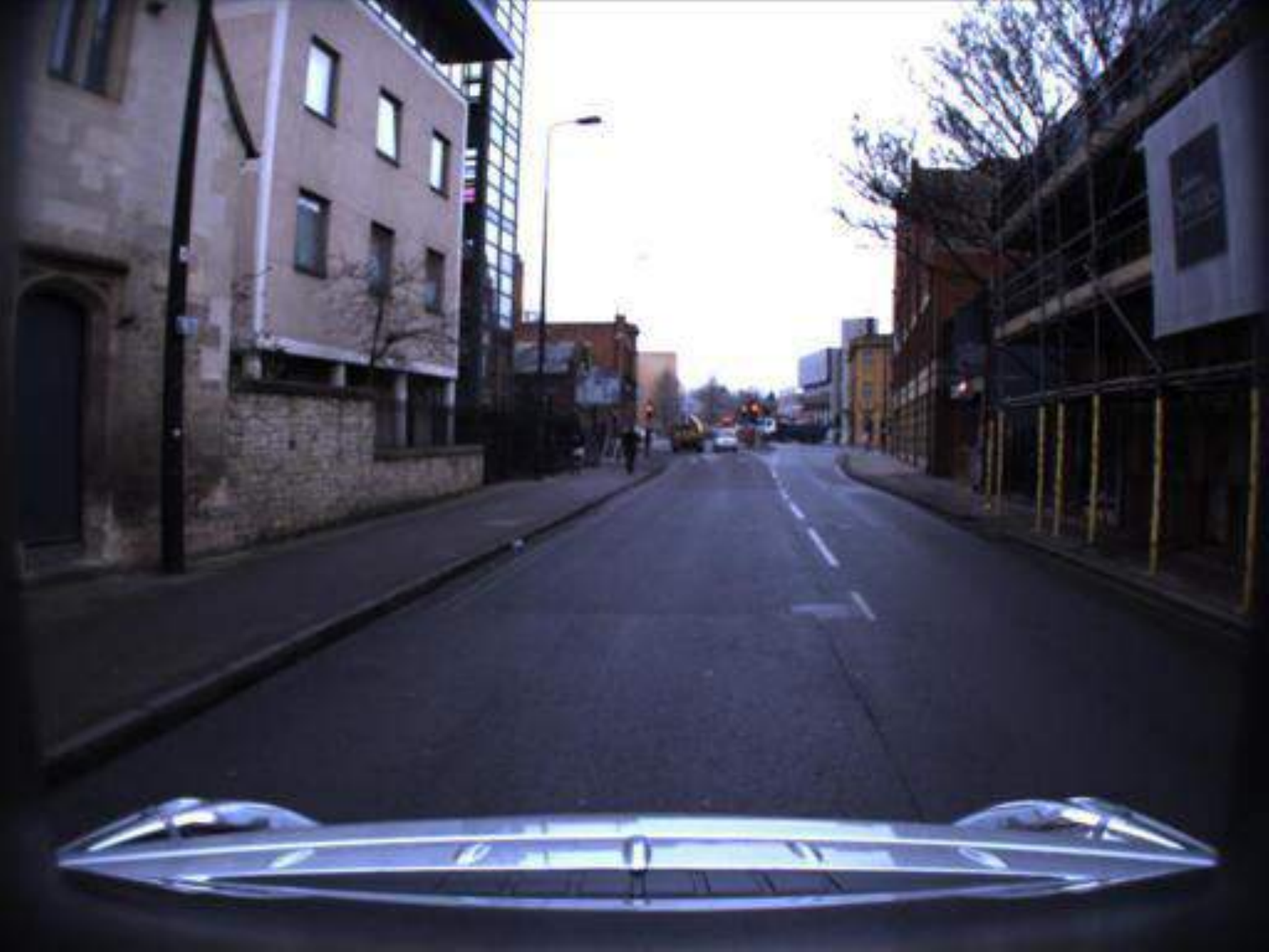} 
        \end{subfigure}
        \begin{subfigure}{\textwidth}
            \includegraphics[width=\textwidth]{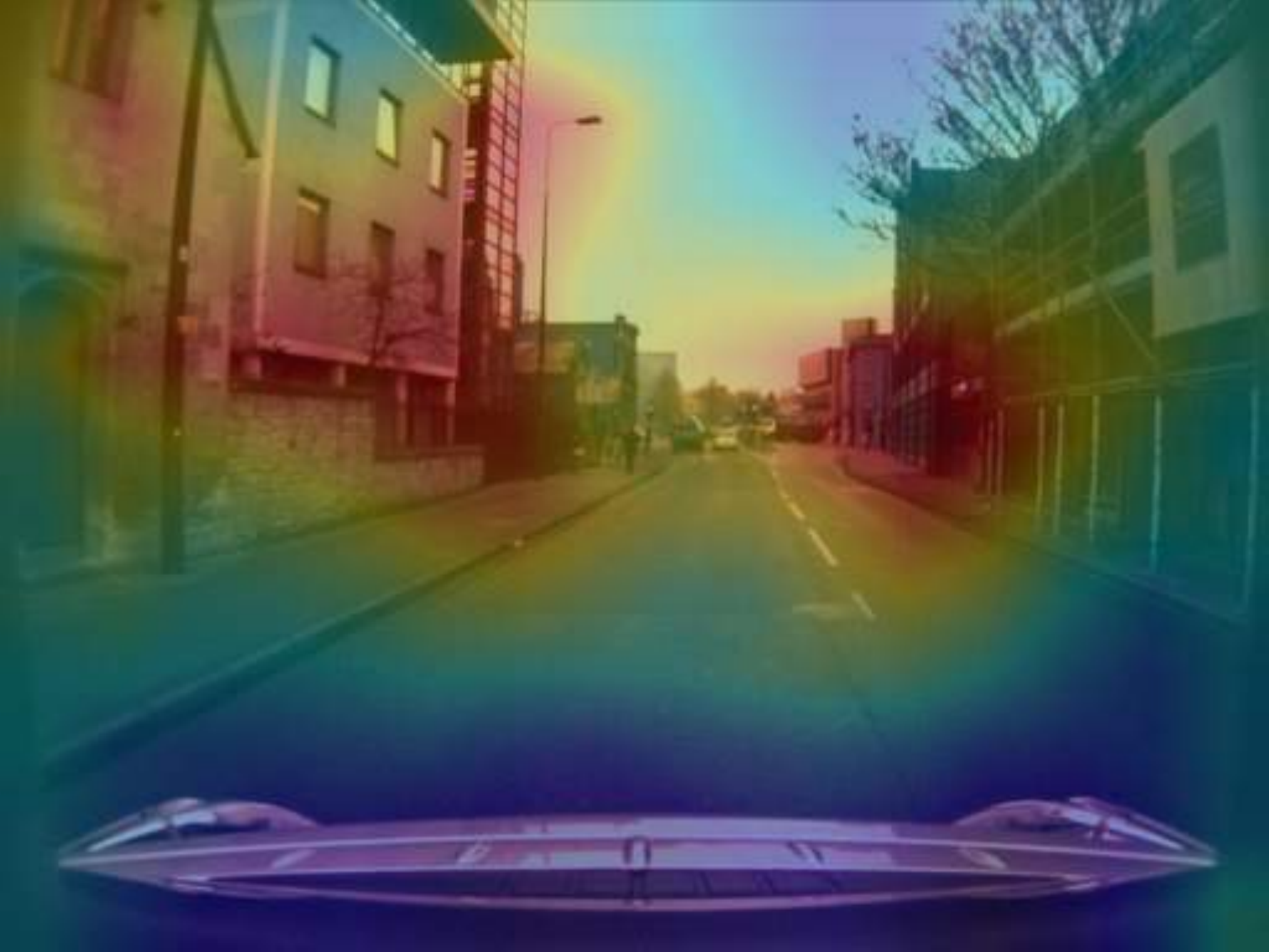} \subcaption{}
        \end{subfigure}
    \end{minipage}
    \caption{Visualization of attention score maps on source domain (a) and target domains (b-f) unseen by the attention module at training time.}
    \label{fig:cam_saliency}
\end{figure}

In our architecture we used the fully connected layer of a CNN pretrained on Places365 \cite{places}, which contains $C=365$ classes, to produce the $AM_c$. The idea stems from the fact that the classes in Places365 \cite{places} (such as house, building, market) are inherently relevant to our task. The images are passed to the whole backbone extracting the local features representation $f$ from the last convolutional layer and producing the $AM_{c_{max}}$ for the category $c_{max}$ with the highest probability $\mathbb{P}$, predicted by the fully connected layer. Then, the features are spatially weighted with the scores calculated before:
\begin{equation}
\begin{split}
    % \mathcal{P}
    f^w &= f \cdot AM_{c_{max}}\\
    c_{max} = c_{i} \mid i =& \argmax_{j}[\mathbb{P}(c_{j})], \forall j \in C
\end{split}
\end{equation}
producing new weighted features $f^w$ with the same dimensions as $f$.\\
We demonstrate that the attention mechanism is useful also for the target images, since the salience regions can help to distinguish also the elements across different domains.
Fig. \ref{fig:cam_saliency} shows the results obtained applying the attention mechanism over all domains at test-time, which shows significant visual results also over target domains unseen by the attention module.

\subsection{Domain adaptation module}
\label{subsec:4.3_grl}
\noindent
In order for the retrieval to work well across domains it is important that the embeddings produced by the attention module are domain agnostic, i.e. they do not encode domain-specific information. 
We achieve this by using a domain discriminator which receives embeddings from the three domains $D_{s}$, $D_{pt}$ and $D_{t}$. The discriminator is composed of two fully connected layers, and its goal is to classify the domain to which the embeddings belong. Just before the discriminator there is a gradient reversal layer (GRL) \cite{grl}, that in the forward pass acts as an identity transform, while in the backward pass multiplies the gradient by -$\lambda$, where $\lambda > 0$. The use of this layer effectively sets up a minimax game strategy, where the discriminator tries to minimize the domain classification loss, that is a cross-entropy loss $\mathcal{L}_{CE}$, while the feature extractor learns to produce domain-invariant embeddings, acting as an adversary to the discriminator.

\subsection{Weakly supervised descriptors aggregation}
\label{subsec:4.3_netvlad}
\noindent
In order to transform the attentive embeddings into vectorized representations of each image we use a NetVLAD \cite{netvlad} layer, arguably the most common descriptor aggregator for VPR \cite{netvlad,kim,wang}.
To use NetVLAD, we first perform K-means clustering over 500 randomly sampled embeddings of images from all 3 domains to compute $K$ centroids. Then, given the embeddings $f^w$ of dimension $D \times H_1 \times W_1$, reshaped with dimensions $D \times R$ where $R = H_1 \times W_1$, the $(j,k)$ element of the VLAD representation $V$ \cite{vlad} is computed as
\begin{equation}
    V(j,k) = \sum_{i=1}^{R} {\frac{e^{- \Vert f^w_{i} - c_{k}\Vert ^{2}}} 
                 {\sum_{k'} e^{- \Vert f^w_{i} - c_{k'}\Vert ^{2}}} \cdot (f^w_{i}(j) - c_k(j))}
\end{equation}
where $f^w_{i}(j)$ and $c_k(j)$ are the $j$-th dimensions of the $i$-th
embedding and $k$-th centroids, respectively; while the fraction is the soft-assignment of descriptor $f^w_{i}$ to centroid $k$-th.
Given the intrinsic nature of VPR data, where the label for each image is represented solely by its position, it is not possible to use standard supervised losses to drive the training, because two photos taken in the same position (therefore with the same label) but opposite directions would depict different locations.
To overcome this, we use a weakly supervised triplet margin loss \cite{netvlad}, which for each query $q$ is defined as
\begin{equation}
\begin{split}
    \mathcal{L}_{triplet} = \sum_y^Y h(\min_i {d^2(F(q), F(p_i^q))} + m \\ - d^2(F(q), F(n_y^q)))
\end{split}
\end{equation}
where $d (\hspace{2pt}\cdot\hspace{2pt},\hspace{2pt}\cdot\hspace{2pt})$ represents the Euclidean distance, $F(x)$ is the features representation for image x, $\{p_i^q\}$ is the set of potential hard positives (images within 10 meters from the query $q$), $\{n_y^q\}$ is the set of $Y$ negatives (further than 25 meters), $h$ is the hinge loss and $m$ is a constant parameter chosen as margin.

\section{Experiments}
\label{sec:experiments}
\noindent
In this section we explain the experimental protocol, focusing on the methods considered for comparisons, the training details for AdAGeo, the experimental results and an ablation study.

\subsection{Comparisons with other methods}
\label{sec:exp_baseline}
\noindent
To compare AdAGeo with other methods, we first consider NetVLAD \cite{netvlad}, arguably the most used and well-established method for visual place recognition. We also compute results with the only other method built for VPR with domain adaptation, by Wang et al. \cite{wang}, which uses an attention module and MK-MMD \cite{mkmmd}, with the code provided by the authors. We used the two variants proposed by the authors, the first one with just the attention mechanism (Wang: Att) and the second one with also the DA branch (Wang: Att+DA). 
Given the lack of other methods for the task, we implement NetVLAD \cite{netvlad} with a GRL \cite{grl} branch, as well as NetVLAD \cite{netvlad} with a DeepCORAL \cite{deepcoral} branch and NetVLAD \cite{netvlad} with an SAFN \cite{SAFN} branch, as SAFN is chosen as the current state of the art for domain adaptation. For SAFN, we compute the features norm from the embeddings produced by the last convolutional layer of the backbone, using the code provided by the authors. For fairness of comparisons, we compare the methods using as backbones AlexNet \cite{alexnet} and ResNet18 \cite{resnet}, both cropped at the last convolutional layer, pretrained on Places365 \cite{places}.

\subsection{Training details}
\label{sec:train_details}
\noindent
The training process is split in two distinct phases, as shown in Fig. \ref{fig:ch3_arch}.
The first phase is tasked with building the pseudo-target dataset using $n^t = 5$ target domain images (Sec . \ref{subsec:4.1_few-shotDDDA}). We adopt the successful architecture of \cite{biost} consisting in two encoders made of two convolutional layers and four residual blocks, and two symmetric decoders made of four residual blocks and two deconvolutional layers. In this phase we use the Adam optimizer with learning rate 0.0002 and batch size 1.
The second phase is tasked with building the embedding for each image, and the domain adaptation task is performed using the same target domain images as in the first phase. The backbone pretrained on Places365 \cite{places} is finetuned from the last two convolutional blocks to the end (both for AlexNet \cite{alexnet} and ResNet18 \cite{resnet}), while the features are extracted at the last convolutional layer, before ReLU, to be passed to the attention and the domain adaptation modules. As optimizer we use Adam with learning rate 0.00001, and for each iteration we use 4 tuples, each consisting of 1 query image, the best positive, and 10 negative samples. 
The negative samples are chosen following the standard described in \cite{netvlad}, in order to increase the likelihood that $\mathcal{L}_{triplet} > 0$, by making sure that each negative is similar enough to the positive.
The two losses are combined as $\mathcal{L}_{triplet} + \alpha \cdot \mathcal{L}_{CE}$ where $\alpha = 0.1$. 
Finally, unlike most domain adaptation methods which train the network for a constant number of epochs, or perform validation and early stopping on the source validation set, we perform validation and early stopping on the generated pseudo-target validation set which, having a similar distribution to the target set, helps to stop the training in an optimal position.

\subsection{Results}
\label{sec:results_sec}
\noindent
All methods are trained on SVOX dataset (Tab. \ref{tab:dataset_counter}).
\begin{figure*}
    \centering
    \subfloat[][\emph{}]{
        \includegraphics[width=0.5\textwidth]{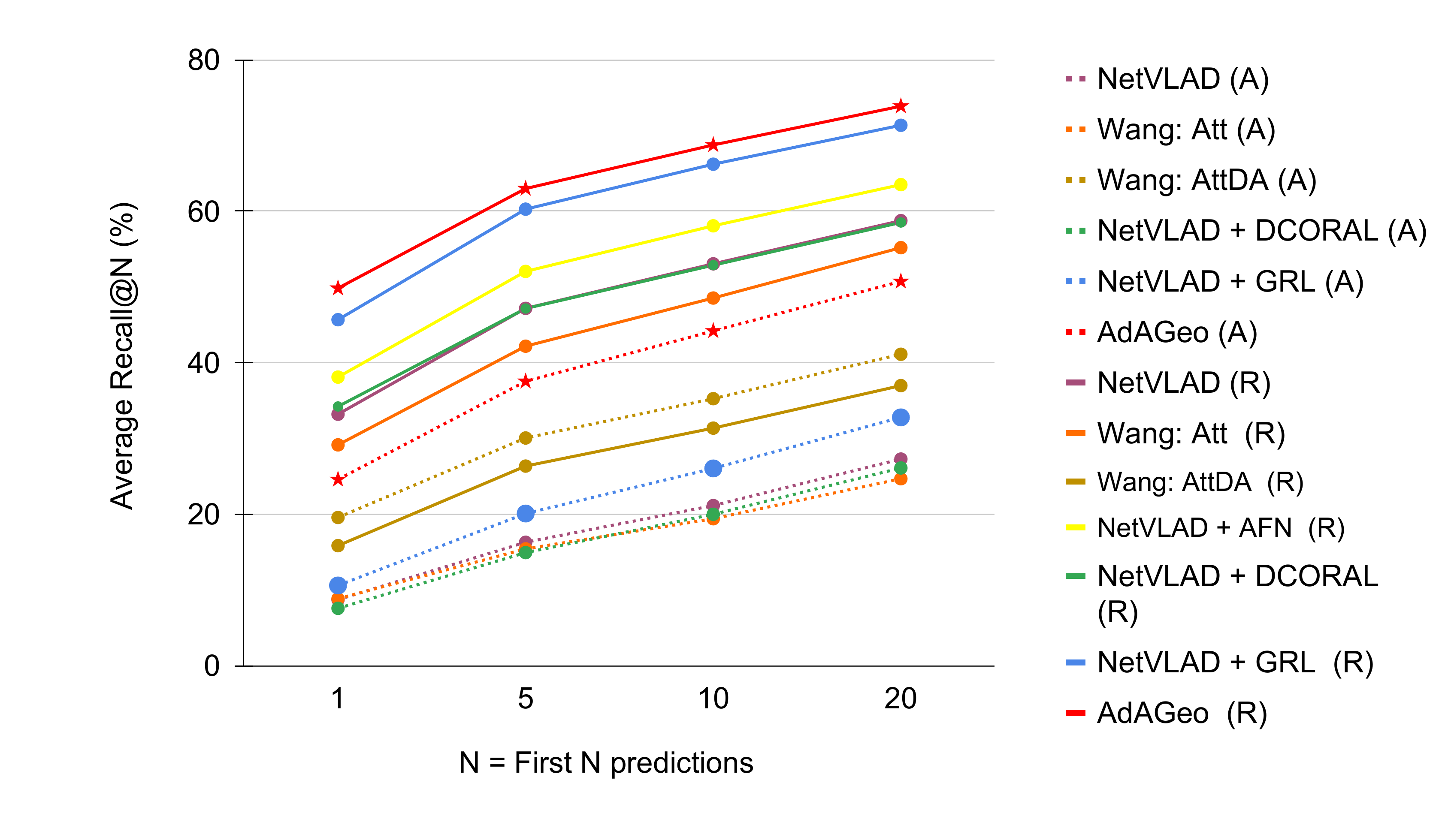}
        \label{fig:exp_npredictions}
    } 
    \subfloat[][\emph{}]{
        \includegraphics[width=0.5\textwidth]{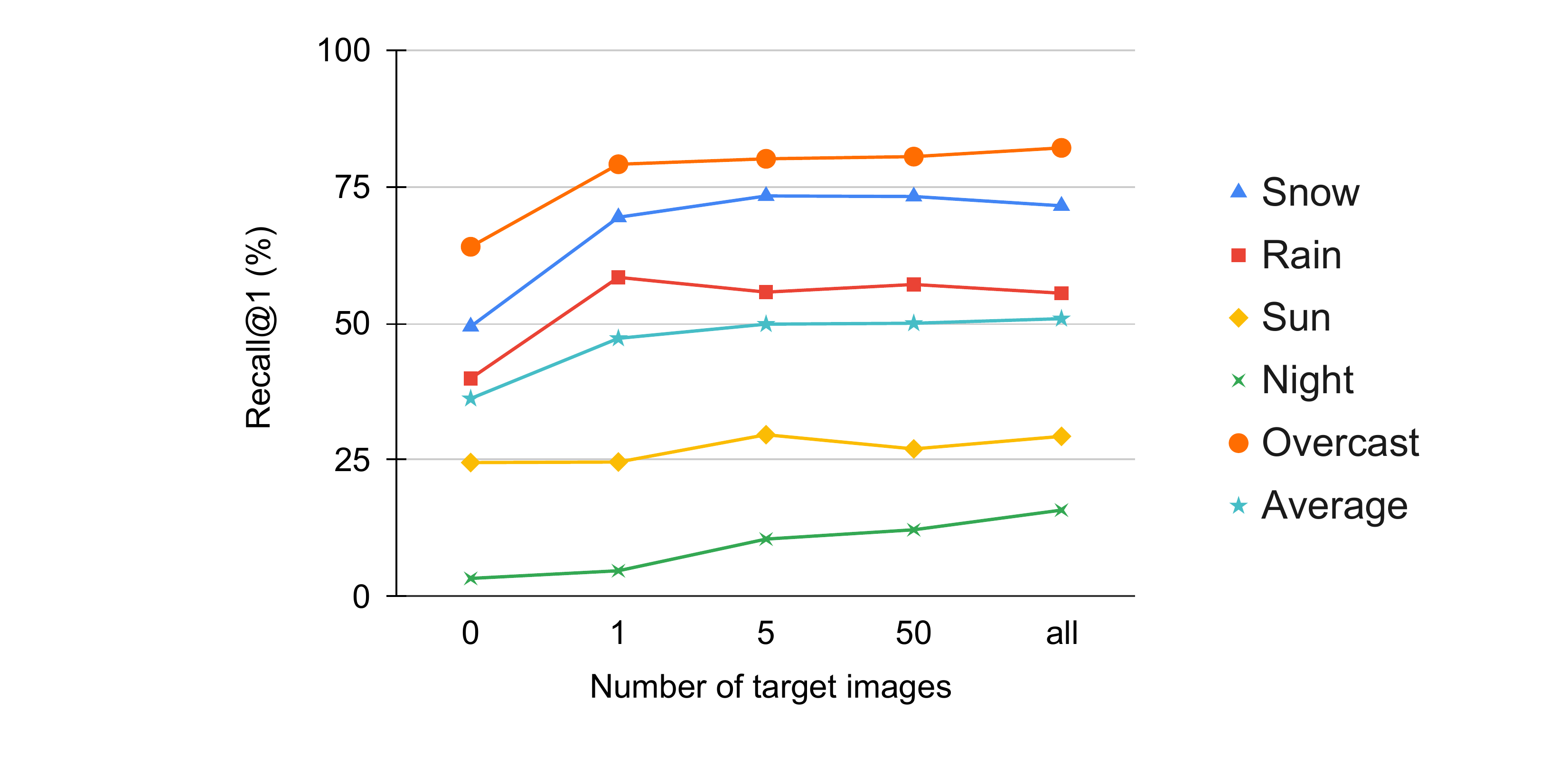}
        \label{fig:exp_ntarget}
    }
    \caption{\textbf{a)} Comparison between all methods, shown as recall@N averaged over the 5 target domains (Average Recall@N). The base CNN encoder is denoted in brackets: (A)lexNet and (R)esNet18. \textbf{b)} Results of experiments with AdAGeo with 0-shots, 1-shot, 5-shots, 50-shots and all-shots. With easier domains (Snow, Rain, Overcast) AdAGeo shows a good improvement in accuracy with just 1 target domain image, while with more challenging domains (Sun, Night) AdAGeo requires a higher amount of images to perform significant improvements.}
    \label{fig:charts}
\end{figure*}
\begin{table*}
    \centering
    \begin{tabular}{lC{0.5cm}C{0.5cm}C{1.45cm}C{1.45cm}C{1.45cm}C{1.45cm}C{1.45cm}C{1cm}} 
        \toprule
        \multicolumn{1}{l}{\multirow{2}{*}{Method}} & \multirow{2}{*}{EN} & \multirow{2}{*}{\#T} & Snow & Rain & Sun & Night & Overcast & \multirow{2}{*}{Avg} \\ 
        \cline{4-8}
        \multicolumn{1}{c}{} &&& R@1 & R@1 & R@1 & R@1 & R@1 & \\
        \hline
        NetVLAD\cite{netvlad}                 & A & 0 &  9.8 \footnotesize{ $\pm$ 2.6} &  9.2 \footnotesize{ $\pm$ 1.5} & 2.0 \footnotesize{ $\pm$ 0.2}  &  0.0 \footnotesize{ $\pm$ 0.0} & 22.5 \footnotesize{ $\pm$ 4.5} &  8.7 \\
        Wang: Att\cite{wang}                  & A & 0 & 11.7 \footnotesize{ $\pm$ 1.5} &  9.8 \footnotesize{ $\pm$ 0.6} & 2.9 \footnotesize{ $\pm$ 0.6}  &  0.2 \footnotesize{ $\pm$ 0.1} & 19.6 \footnotesize{ $\pm$ 2.0} &  8.8 \\
        Wang: Att+DA\cite{wang}               & A & all & 28.7 \footnotesize{ $\pm$ 1.5} & 20.6 \footnotesize{ $\pm$ 2.5} & 6.4 \footnotesize{ $\pm$ 0.2}  &  0.8 \footnotesize{ $\pm$ 0.2} & 41.3 \footnotesize{ $\pm$ 1.0}  &  19.6 \\
        NetVLAD\cite{netvlad}+DCORAL\cite{deepcoral}   & A & all &  9.6 \footnotesize{ $\pm$ 1.1} &  8.7 \footnotesize{ $\pm$ 1.3} & 2.2 \footnotesize{ $\pm$ 0.3}  &  0.1 \footnotesize{ $\pm$ 0.1} & 17.2 \footnotesize{ $\pm$ 2.0} &  7.6 \\
        NetVLAD\cite{netvlad}+GRL\cite{grl}   & A & all &  12.4 \footnotesize{ $\pm$ 3.4} &  9.2 \footnotesize{ $\pm$ 1.8} & 3.2 \footnotesize{ $\pm$ 0.3}  &  0.1 \footnotesize{ $\pm$ 0.2} & 28.0 \footnotesize{ $\pm$ 2.4} &  10.6 \\
        AdAGeo (ours)                         & A & 5 &  \textbf{34.9} \footnotesize{ $\pm$ 2.2} &  \textbf{26.4} \footnotesize{ $\pm$ 3.3} & \textbf{10.0} \footnotesize{ $\pm$ 0.1} &  \textbf{1.7} \footnotesize{ $\pm$ 0.4} & \textbf{49.9} \footnotesize{ $\pm$ 1.9}  & \textbf{24.6} \\
        \hline
        NetVLAD\cite{netvlad}                 & R & 0 & 50.1 \footnotesize{$\pm$ 1.3} & 36.5 \footnotesize{$\pm$ 0.6} & 17.7 \footnotesize{$\pm$ 0.9} &  1.6 \footnotesize{$\pm$ 0.4} & 60.0 \footnotesize{$\pm$ 0.7} & 33.2 \\
        Wang: Att\cite{wang}                  & R & 0 & 47.2 \footnotesize{$\pm$ 5.0}  & 28.1 \footnotesize{$\pm$ 3.3}  & 13.5 \footnotesize{$\pm$ 1.9}  &  1.3 \footnotesize{$\pm$ 1.4}  & 55.7 \footnotesize{$\pm$ 4.6}  & 29.2 \\
        Wang: Att+DA\cite{wang}               & R & all & 23.8 \footnotesize{$\pm$ 6.2} & 11.2 \footnotesize{$\pm$ 1.4} & 5.7 \footnotesize{$\pm$ 0.5}  &  0.9 \footnotesize{$\pm$ 0.5} & 37.6 \footnotesize{$\pm$ 8.2} & 15.8 \\
        NetVLAD\cite{netvlad}+SAFN\cite{SAFN} & R & all & 57.3 \footnotesize{$\pm$ 2.5} & 43.6 \footnotesize{$\pm$ 0.4} & 19.1 \footnotesize{$\pm$ 2.0} &  2.2 \footnotesize{$\pm$ 0.7} & 68.3 \footnotesize{$\pm$ 1.2} & 38.1 \\
        NetVLAD\cite{netvlad}+DCORAL\cite{deepcoral} & R & all & 60.2 \footnotesize{$\pm$ 2.0}  & 33.5 \footnotesize{$\pm$ 1.1}  & 14.1 \footnotesize{$\pm$ 0.6} &  2.1 \footnotesize{$\pm$ 0.8} & 61.2 \footnotesize{$\pm$ 3.6} & 34.2 \\
        NetVLAD\cite{netvlad}+GRL\cite{grl}   & R & all & 68.9 \footnotesize{$\pm$ 2.5} & 50.9 \footnotesize{$\pm$ 2.0} & 27.1 \footnotesize{$\pm$ 4.8} &  4.6 \footnotesize{$\pm$ 1.2} & 76.9 \footnotesize{$\pm$ 0.7} & 45.7 \\
        AdAGeo (ours)                         & R & 5 & \textbf{73.3} \footnotesize{$\pm$ 2.2} & \textbf{55.7} \footnotesize{$\pm$ 1.8} & \textbf{29.6} \footnotesize{$\pm$ 1.0} & \textbf{10.5} \footnotesize{$\pm$ 1.9} & \textbf{80.1} \footnotesize{$\pm$ 1.5} & \textbf{49.8} \\
        \bottomrule
    \end{tabular}
    \caption{Comparison between all methods, shown as recall@1 (R@1) on each target domain. Column EN stands for the encoder used: AlexNet (A) or ResNet18 (R). \#T shows the number of target images used at training time. Snow, Rain, Sun, Night and Overcast are the 5 target domains of the SVOX+RobotCar dataset. The last column shows the average recall@1 across all domains.}
    \label{tab_results}
\end{table*}
\begin{table*}
    \centering
    \begin{tabular}{lC{1.25cm}C{1.25cm}C{1.25cm}C{1.25cm}C{1.25cm}C{1.25cm}} 
        \toprule
        \multicolumn{1}{l}{\multirow{2}{*}{Method}} & Snow & Rain & Sun & Night & Overcast & \multirow{2}{*}{Avg} \\ 
        \cline{2-6}
                          & R@1  & R@1  & R@1  & R@1 & R@1  & \\ 
        \hline
        Baseline          & 50.1 & 36.5 & 17.7 & 1.6 & 60.0 & 33.2 \\
        Baseline+DDDA     & 61.3 & 45.3 & 23.3 & 6.1 & 71.1 & 41.4 \\
        Baseline+Att      & 49.4 & 39.9 & 24.5 & 3.3 & 64.0 & 36.2 \\
        Baseline+DA       & 65.3 & 49.7 & 25.4 & 6.0 & 75.2 & 44.3 \\
        Baseline+DDDA+Att & 66.6 & 54.5 & 27.3 & 5.5 & 72.2 & 45.2 \\
        Baseline+DDDA+DA  & 67.2 & 51.5 & 24.8 & 9.4 & 78.4 & 46.3 \\
        Baseline+Att+DA   & 66.0 & 49.1 & 24.8 & 3.2 & 76.1 & 43.8 \\ 
        \hline
        AdAGeo  & \textbf{73.3} & \textbf{55.7} & \textbf{29.6} & \textbf{10.5} & \textbf{80.1} & \textbf{49.8} \\
        \bottomrule
    \end{tabular}
    \caption{Ablation table of our proposed framework on the SVOX+RobotCar dataset in a 5-shot setting with ResNet18 as encoder. R@1 = recall@1, DDDA = Domain-Driven Data Augmentation, Att = Attention layer and DA = Domain adaptation layer.}
    \label{tab_ablation}
\end{table*}
For methods which use domain adaptation (DA), we used the whole unlabeled target train set from Oxford RobotCar \cite{robotcar} (around 800 images, depending on the domain, see Tab. \ref{tab:dataset_counter}) for the DA task.
For our architecture, we only used 5 images from the unlabeled target set for DA, simulating a five-shots scenario.
Testing is then performed using the test gallery from SVOX and the test queries from Oxford RobotCar \cite{robotcar}. For methods with DA, trainings are performed separately for each of the 5 target domains (Snow, Rain, Sun, Night and Overcast).
As evaluation metric, we use the percentage of correctly localized queries within the first N predictions, known as recall@N, as standard practice for place recognition \cite{netvlad,wang,kim,largescale,multiscale,apanet}. A query is deemed correctly localized if at least one of the top N retrieved gallery images is within 25 meters from the ground truth position of the query.
Results for each method over each domain are shown in Tab. \ref{tab_results}. Our AdAGeo framework outperforms all other approaches with both AlexNet and ResNet18 encoders while using two orders of magnitude less target domain images, which verifies the effectiveness of our method. Moreover, AdAGeo presents good results with both encoders, showing the stability of the framework, while other methods are highly dependent on the architecture of the features extractor. More comparisons of each method are shown in Fig. \ref{fig:exp_npredictions}. The supplementary material provides an additional qualitative comparison between our method and the best baseline by visualizing some retrieval results.\\
For fairness of comparison, we ran all experiments 3 times in a fully deterministic environment, with seeds 0, 1 and 2, and we present the mean over the 3 runs.

\subsection{Ablation study}
\label{sec:ablation_sec}
\noindent
We evaluate the components of our method by conducting an extensive ablation study over each target domain of SVOX+RobotCar. The results are shown in Tab. \ref{tab_ablation}, where all experiments have been run in a 5-shot environment (except for the experiments where the target domain is not used) and all the modules combination are tried. As baseline, we use a ResNet18 encoder (cropped at the last convolutional layer) followed by a NetVLAD \cite{netvlad} descriptor aggregator. Then, each component is added to the baseline: Baseline + Domain-driven data augmentation module (DDDA), Baseline + Attention module (Att), Baseline + Domain adaptation module (DA) and all their combinations (Baseline+DDDA+Att, Baseline+DDDA+DA, Baseline+Att+DA) until the entire AdAGeo architecture (Baseline+DDDA+Att+DA). As shown in Tab. \ref{tab_ablation}, each module produces an improvement w.r.t. the baseline. The ablation study also proves that the modules are orthogonal to each other, giving consistent improvements when used alone as when used together. In particular, the attention module yields a 3\% improvement on the baseline, and 3.5\% on the final model, although it does not see the target domain at training time. Finally, the three modules together show an improvement of more than 16\% on average over the baseline.

\section{Conclusions}
\label{sec:conclusions}
\noindent
In this work we propose AdAGeo, a method to tackle the problem of cross-domain visual place recognition using only few unlabeled target images. The key improvements over previous architectures are due to an attention mechanism, and two orthogonal domain adaptation techniques. 
We extensively show the robustness of AdAGeo, especially when only few target images are available for domain adaptation at training time, being able to outperform current state of the art with two orders of magnitude less target images.
Moreover, we propose a new dataset, called SVOX, which, extends Oxford RobotCar and can be used as a large scale multi-domain dataset for visual place recognition, presenting a realistic scenario for future research on the field.

{\small
\bibliographystyle{ieee_fullname}
\bibliography{bibliography}
}

\section{Appendix}
\paragraph{Additional dataset details}
The main idea behind SVOX is to build a dataset that contains the RobotCar dataset \cite{robotcar}, in order to test the accuracy of cross-domain visual geolocalization methods.
To create the dataset we downloaded images from Google Street View, which provides 360$^{\circ}$ equirectangular panoramas at various resolutions.
From each panorama we then cropped two rectangles at opposite sides, corresponding to the front and rear view of the car.

The original resolution of the images from the RobotCar dataset \cite{robotcar} is 1280x960, and we resized them to 512x384, keeping the original ratio of 4:3. We resized the images cropped from Google Street View panoramas to the same size, again keeping the same ratio.

\begin{figure*}
    \centering
    \begin{minipage}{.24\textwidth}
        \begin{subfigure}{\textwidth}
            \includegraphics[width=\textwidth]{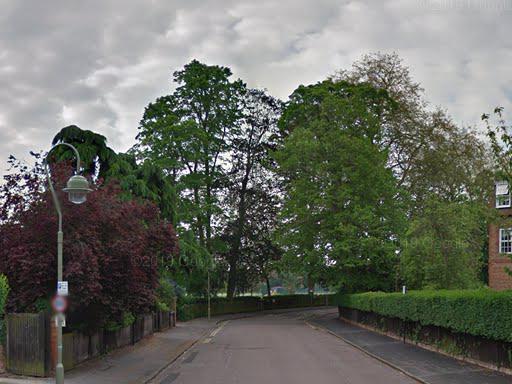}
        \end{subfigure}
        \begin{subfigure}{\textwidth}
            \includegraphics[width=\textwidth]{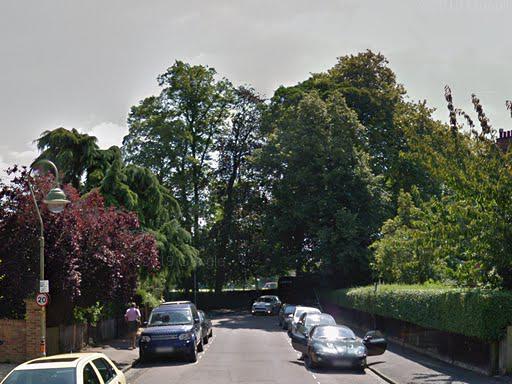} 
            \subcaption{}
        \end{subfigure}
    \end{minipage}
    \begin{minipage}{.24\textwidth}
        \begin{subfigure}{\textwidth}
            \includegraphics[width=\textwidth]{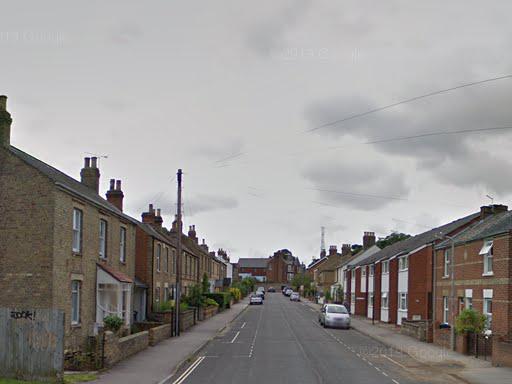} 
        \end{subfigure}
        \begin{subfigure}{\textwidth}
            \includegraphics[width=\textwidth]{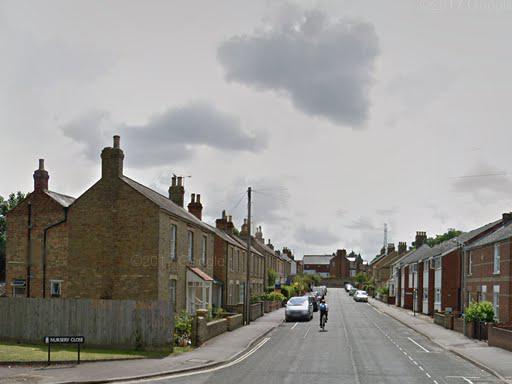} 
            \subcaption{}
        \end{subfigure}
    \end{minipage}
    \begin{minipage}{.24\textwidth}
        \begin{subfigure}{\textwidth}
            \includegraphics[width=\textwidth]{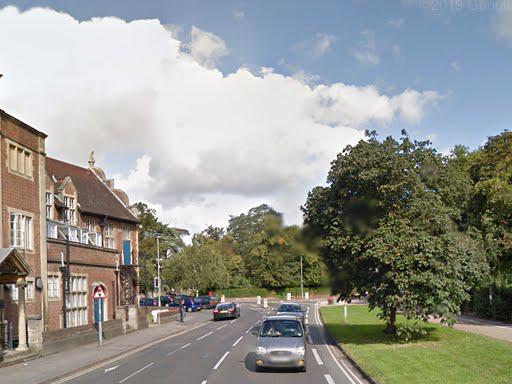} 
        \end{subfigure}
        \begin{subfigure}{\textwidth}
            \includegraphics[width=\textwidth]{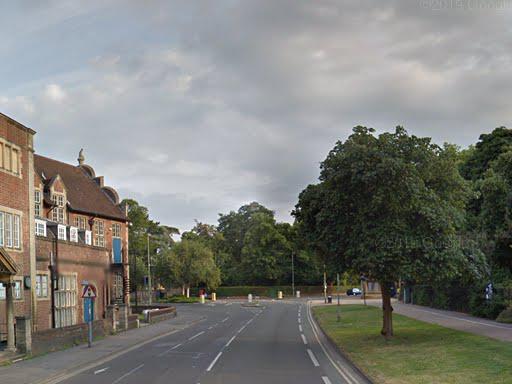} 
            \subcaption{}
        \end{subfigure}
    \end{minipage}
    \begin{minipage}{.24\textwidth}
        \begin{subfigure}{\textwidth}
            \includegraphics[width=\textwidth]{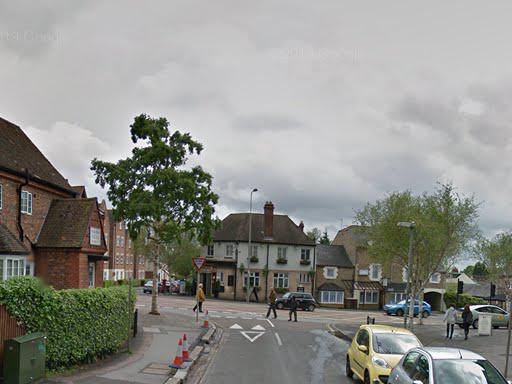} 
        \end{subfigure}
        \begin{subfigure}{\textwidth}
            \includegraphics[width=\textwidth]{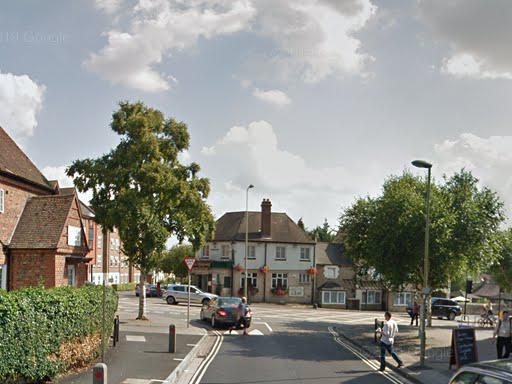} 
            \subcaption{}
        \end{subfigure}
    \end{minipage}
    \caption{Examples of Oxford places at different times by means of Google Time Machine API. On the top row there are the images from 2012 used as gallery set, while on the bottom row there are the images from 2014 used as query set.}
    \label{fig:timemachine}
\end{figure*}

Thanks to the Google Street View Time Machine we are able to download panoramas taken in the same location in different years. We chose to use images from the years of 2012 and 2014 as gallery and queries respectively, as these are the years with most panoramas in the Oxford area. Moreover, using gallery and queries taken in different years helps to ensure that methods that achieve accuracy must focus on long-term elements, instead of short-term or changing elements such as vegetation or scaffolding. The RobotCar dataset \cite{robotcar} was collected between 2014 and 2015, ensuring that the queries from RobotCar \cite{robotcar} are at least two years apart from the SVOX gallery. Some examples are shown in Fig. \ref{fig:timemachine}.

To build SVOX we chose a geographical area that would enclose the whole urban part of the city of Oxford. We then removed by hand images taken in the countryside, given the lack of buildings that are crucial to the geolocalization process. Moreover, we removed queries (from both SVOX and RobotCar \cite{robotcar}) which do not have a positive image within gallery, i.e. and image within 25 meters of distance. Finally we split SVOX in train, validation and test sets. As shown in Fig. 1 of the main paper, the RobotCar dataset \cite{robotcar} is included only in the train and test set, as it is intended to be used only as an unlabeled target dataset for domain adaptation, therefore not requiring a validation set.

\begin{figure*}
    \centering
    \begin{subfigure}{0.9\textwidth}
        \includegraphics[width=\textwidth]{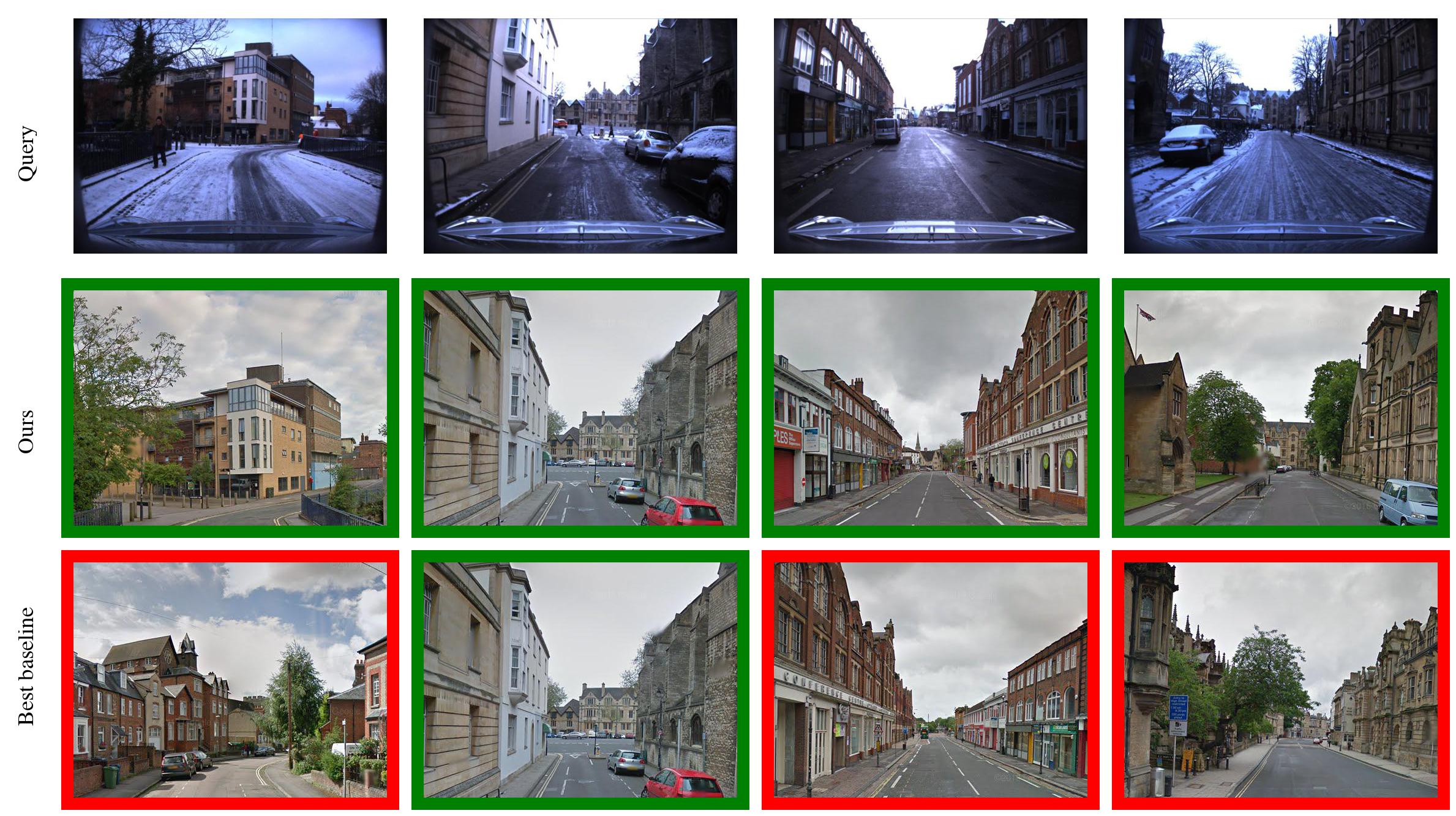}
    \end{subfigure}
    \caption{Comparison between our method and the best baseline, showing the top1 images retrieved for the target scenario Snow. The images with green border correspond with the ground truth, while the ones with a red border are wrong predictions.}
    \label{fig:preds_1}
\end{figure*}

\begin{figure*}
    \centering
    \begin{subfigure}{0.9\textwidth}
        \includegraphics[width=\textwidth]{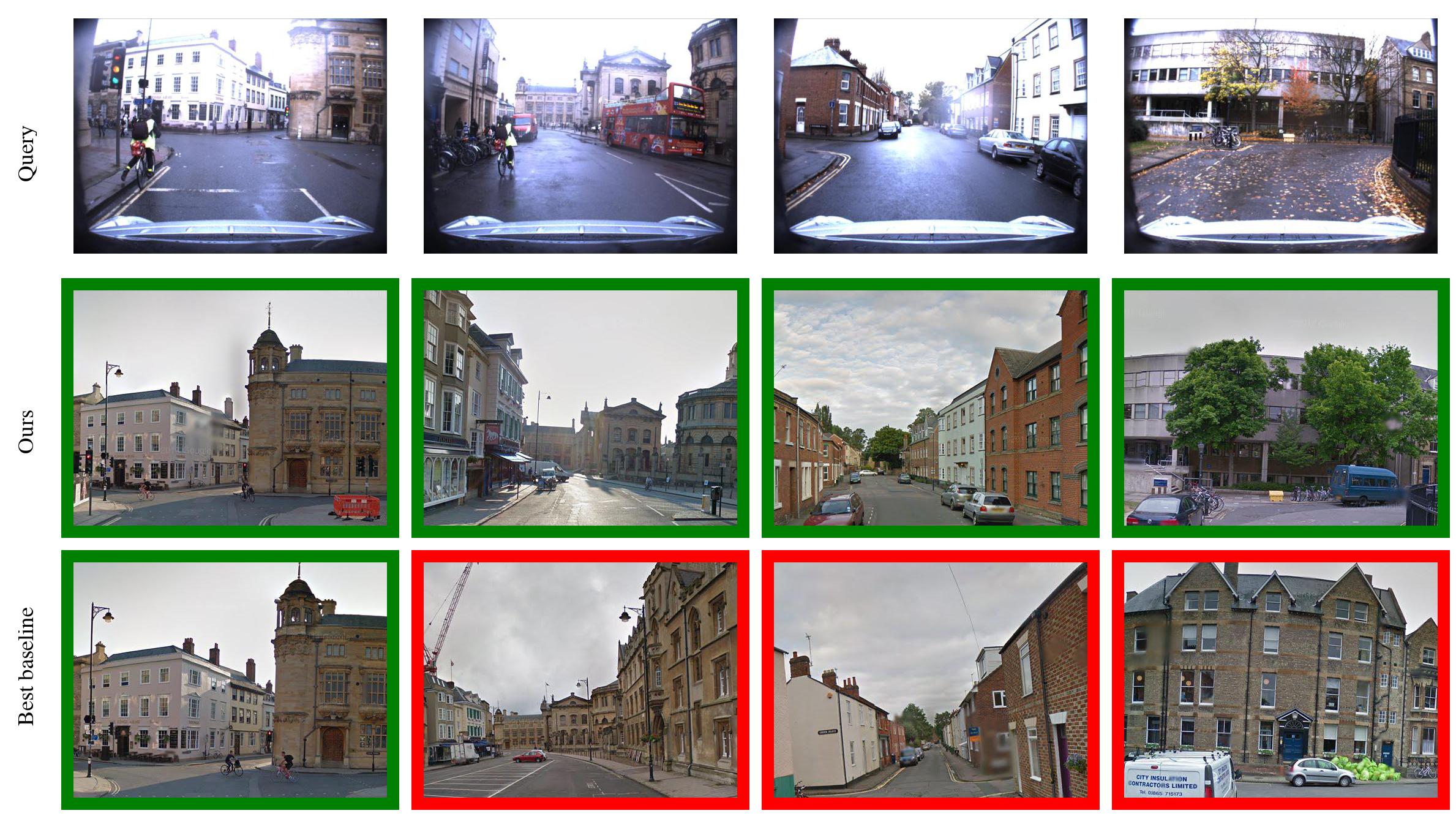}
        \caption{}
    \end{subfigure}
    \vspace{6pt}
    \hrule
    \vspace{6pt}
    \begin{subfigure}{0.9\textwidth}
        \includegraphics[width=\textwidth]{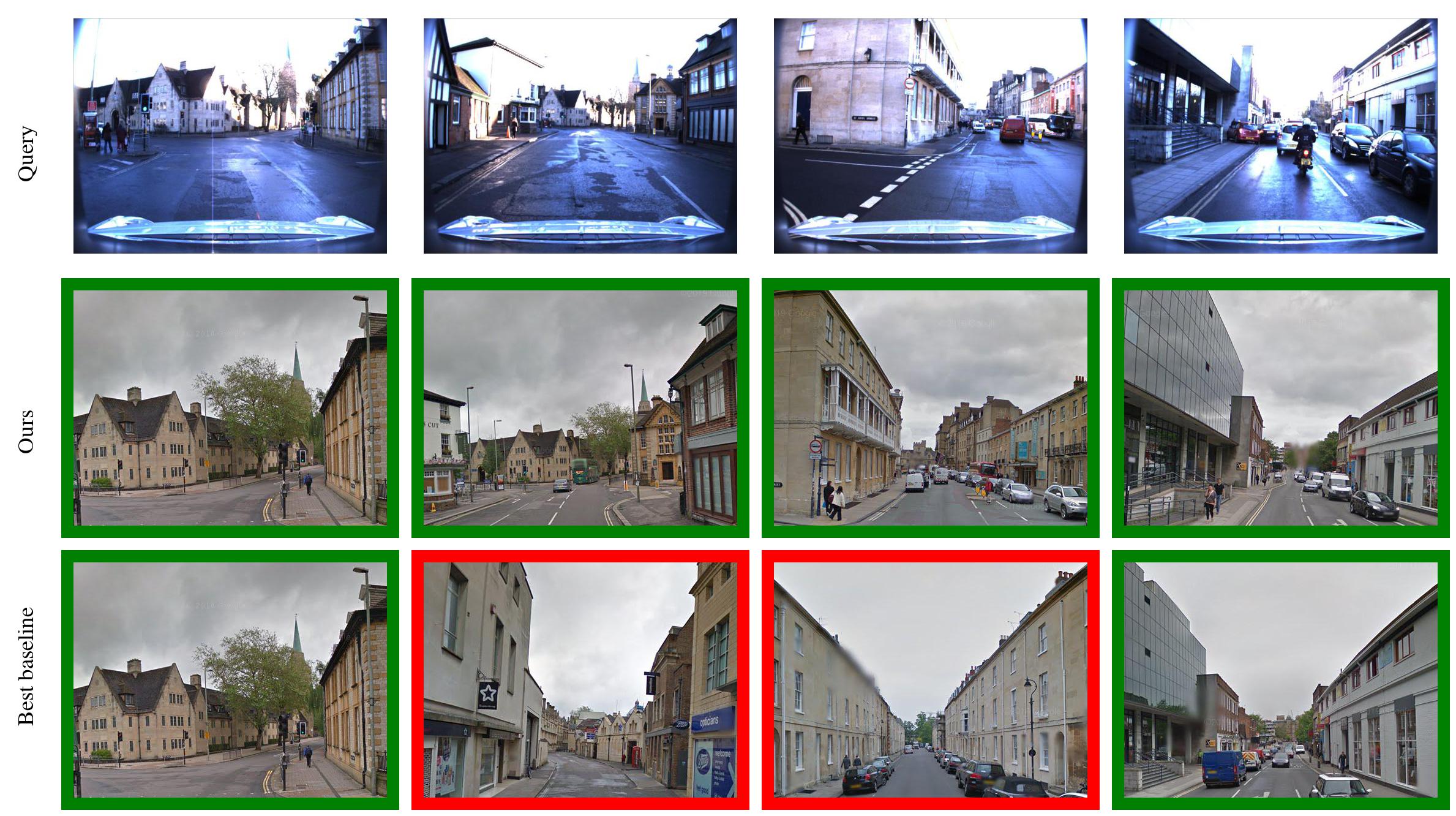}
        \caption{}
    \end{subfigure}
    \caption{Comparison between our method and the best baseline, showing the top1 images retrieved for the target scenarios Rain (a) and Sun (b). The images with green border correspond with the ground truth, while the ones with a red border are wrong predictions.}
    \label{fig:preds_2-3}
\end{figure*}

\begin{figure*}
    \centering
    \begin{subfigure}{0.9\textwidth}
        \centering
        \includegraphics[width=\textwidth]{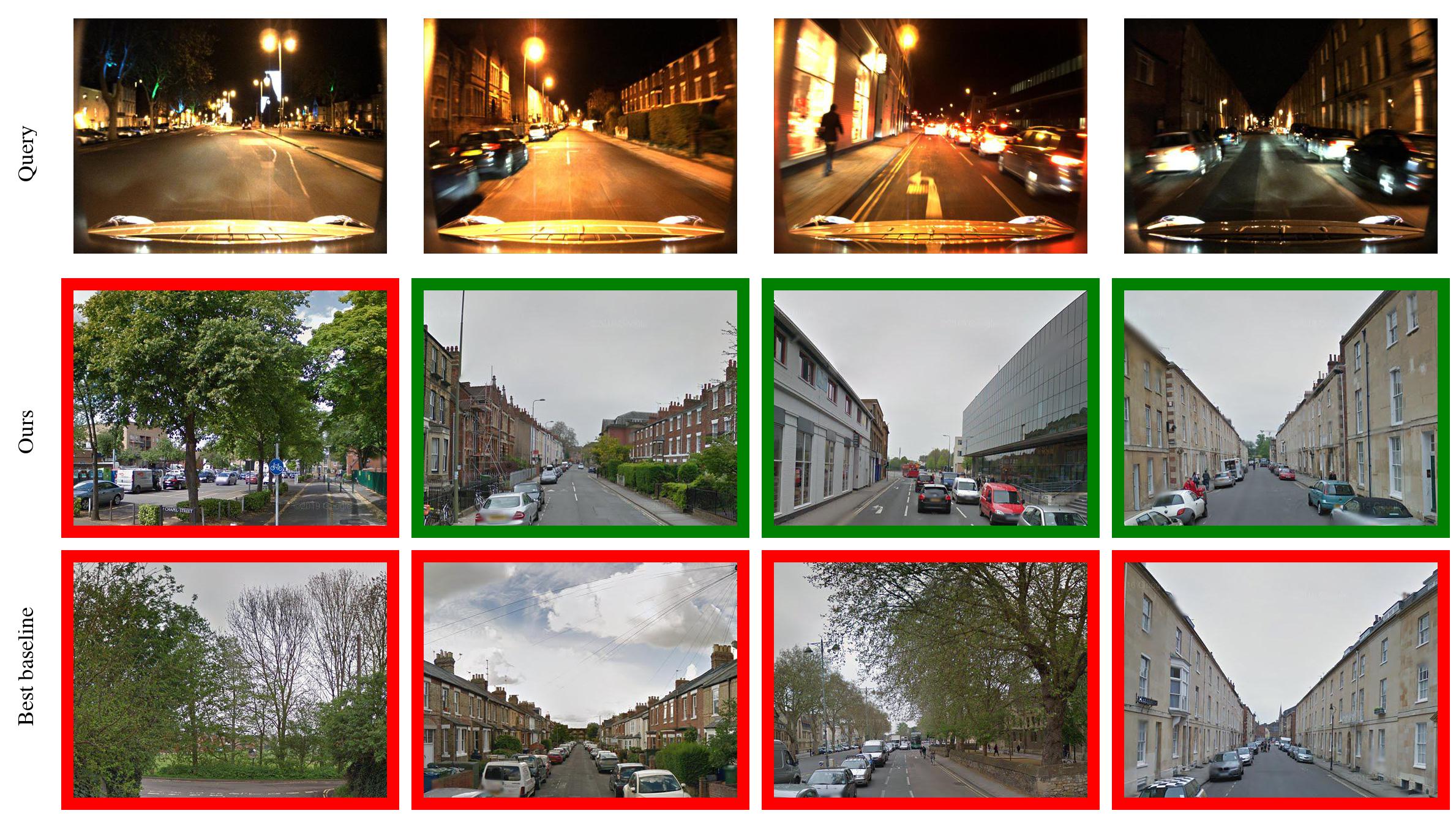}
        \caption{}
    \end{subfigure}
    \vspace{6pt}
    \hrule
    \vspace{6pt}
    \begin{subfigure}{0.9\textwidth}
        \includegraphics[width=\textwidth]{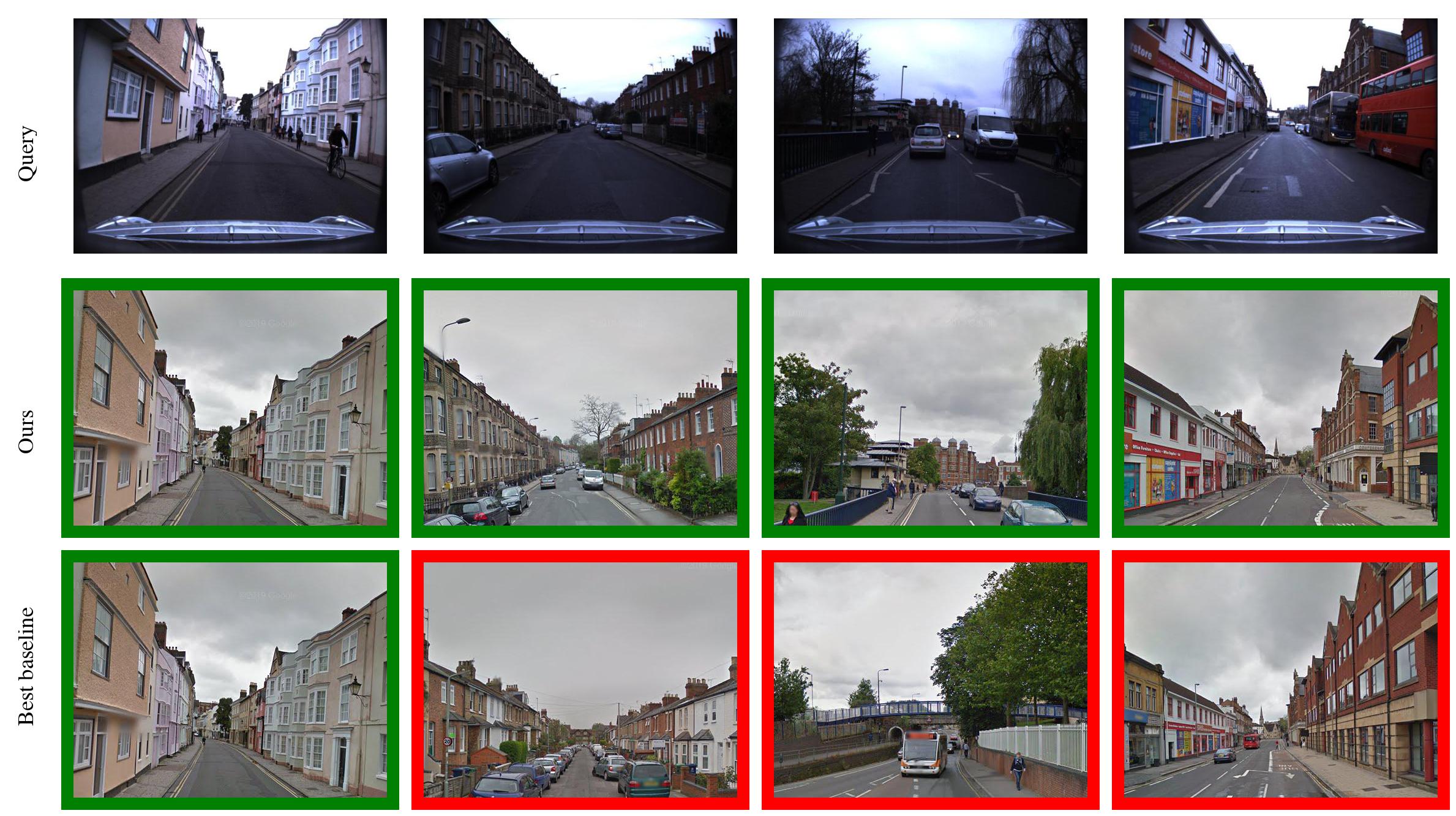}
        \caption{}
    \end{subfigure}
    \caption{Comparison between our method and the best baseline, showing the top1 images retrieved for the target scenarios Night (a) and Overcast (b). The images with green border correspond with the ground truth, while the ones with a red border are wrong predictions.}
    \label{fig:preds_4-5}
\end{figure*}

\paragraph{Qualitative results}
%Qui mostriamo dei risultati visivi, per ogni scenario, confrontando il nostro metodo, con la migliore baseline (NetVLAD + GRL).
In Figs. \ref{fig:preds_1}, \ref{fig:preds_2-3} and \ref{fig:preds_4-5} we show for each target scenario of RobotCar \cite{robotcar}, some visualizations of top1 images retrieved by our method (AdAGeo) versus the best baseline (NetVLAD \cite{netvlad} + GRL \cite{grl}), which are trained and tested with the ResNet18 \cite{resnet} as encoder.

\end{document}